\pdfoutput=1

\documentclass[11pt]{article}

\usepackage[preprint]{acl}

\usepackage{times}
\usepackage{latexsym}

\usepackage[T1]{fontenc}

\usepackage[utf8]{inputenc}

\usepackage{microtype}

\usepackage{inconsolata}

\usepackage{graphicx}

\title{From Data to Insights: Exploring Program-of-Thoughts Prompting for Chart Summarization}

\author{Yutong Qu \\
  Adelaide University \\ Adelaide, SA, Australia \\
  \texttt{yutong.qu@adelaide.edu.au} \\\And
  Wei Zhang \\
  Adelaide University \\ Adelaide, SA, Australia \\
  \texttt{wei.e.zhang@adelaide.edu.au} \\
}

\usepackage{minted}
\usepackage{xcolor}
\usepackage{listings}
\usepackage{wrapfig}
\usepackage{xparse}
\usepackage{graphicx}   
\usepackage{booktabs}   
\usepackage{multirow}   
\NewDocumentCommand{\codeword}{v}{%
\texttt{\textcolor{blue}{#1}}%
}
\usepackage{hyperref}
\usepackage[normalem]{ulem}

\begin{document}
\maketitle

\begin{abstract}
Charts play a critical role in conveying numerical data insights through structured visual representations. However, semantic visual understanding and numerical reasoning requirements hinder the accurate description of charts, interpreting a challenging task in chart summarization.
Despite recent advancements in visual language models (VLMs), approaches lack robust mechanisms for verifying statistical fact correctness and are computationally heavy.
To address this gap, this paper explores a strategy of using zero-shot learning to motivate the lightweight VLMs to perform computational reasoning, via Python programs as intermediaries to derive valid summary statistics for chart understanding.
Specifically, we introduce a novel chart-to-dictionary auxiliary task, offering a more flexible representation compared to traditional chart-to-table methods, making it particularly well-suited for integration with the Program-of-Thought (PoT) strategy.
Experimental results demonstrate our strategy performs on par with existing chart summarization methods across semantic and factual metrics.
Code is available on \url{https://anonymous.4open.science/r/ZeroShot-PoT-C2T-5A6B}.

\end{abstract}

\section{Introduction}




\begin{figure}[h]
  \includegraphics[width=\linewidth]{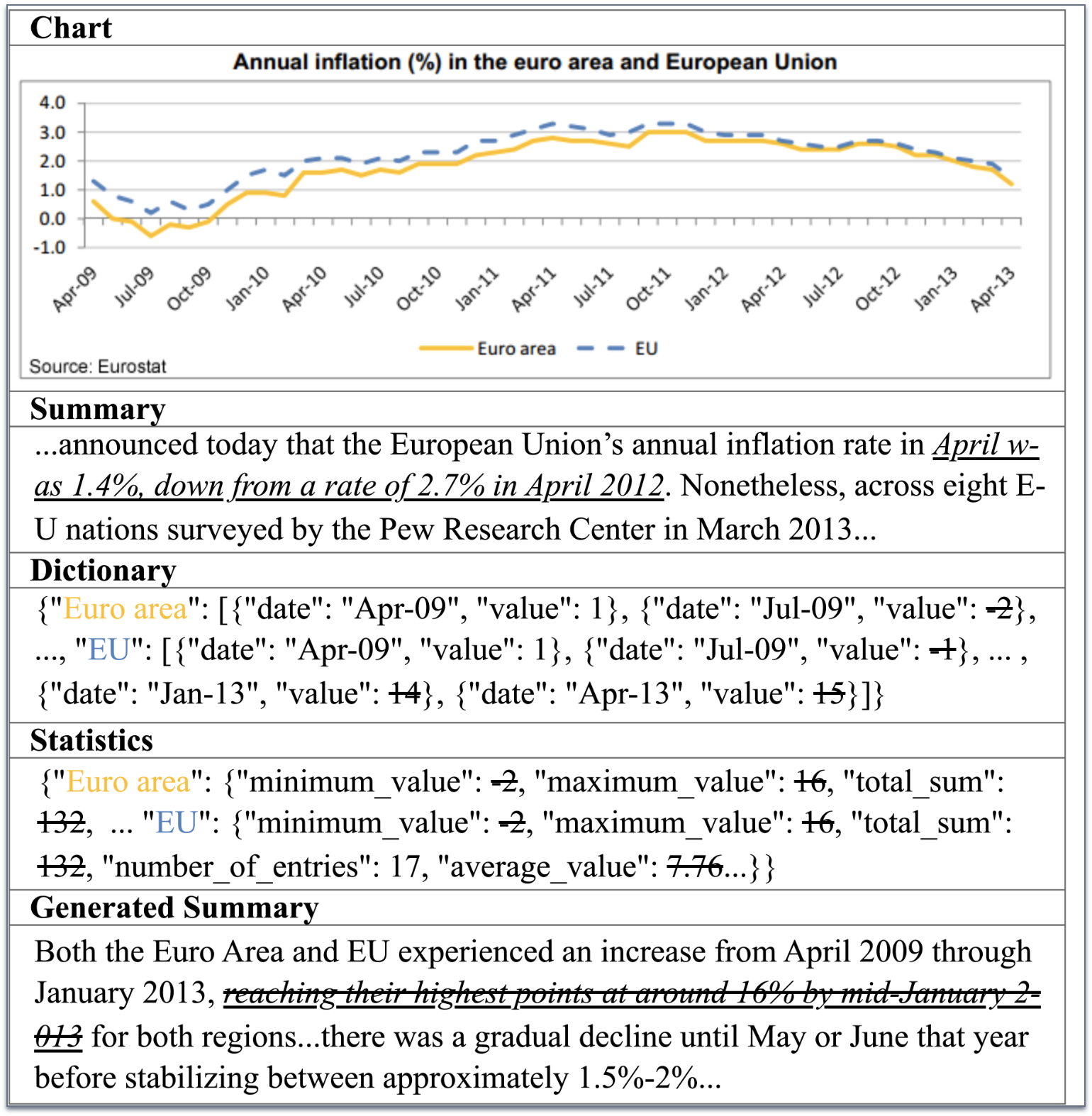}
  \caption{Example of a chart in Pew dataset with its representations in Python dictionary and statistics. \underline{\textit{Italic}} indicates L2/L3 content in chart summarization. \sout{Strikeout} indicates hallucination errors and error-inducing tokens.}
  \label{fig:c2s_diagram}
\end{figure}


With the rising demand for visualizing quantitative data, the growing adoption of digital media has played a role in the rapid growth of data visualization, which has led to the task of automatic chart understanding, information extraction, and summarization, critical areas of research \cite{huang2024pixels, zhang2024tinychart, choi2025end}.
%
%
Recent advancements in Visual Language Models (VLMs) have shown promise in this area \cite{masry2023unichart, han2023chartllama, ko2024chartllm, masry2024chartinstruct, meng2024chartassistant, zhang2024tinychart, liu2024chartthinker}; however, existing methods still struggle with achieving high-quality summaries, especially for L2/L3 content - which is identified as statistics and relations (e.g., min, max) / perceptual and cognitive phenomena (e.g., trends) \cite{lundgard20224levelcontent,kantharaj2022chart,tang2023vistext}, as shown in Figure~\ref{fig:c2s_diagram}.
The challenge is around the highly inconsistent matching between the generated summary and the chart's actual data content, which yields factual inconsistencies and hallucinations. This is either due to failing to parse the text in the chart or to demarcate the numerical value of the visualized data. Additionally, with semantic parsing of the chart elements, VLMs struggle at performing complex reasoning about chart patterns and incorporating statistical reasoning with chart elements \cite{liu2024chartthinker}.
Despite general challenges, although current VLM-based chart understanding methods have shown a certain level of performance, they still face two main challenges:
%
%
(1) Existing implementations are fine-tuned or pre-trained specifically on chart-related instruction data. While this alignment between the vision encoder and language decoder enhances generalization performance, such training processes introduce significant computational overhead, making them resource-intensive and challenging under computational constraints;
%
(2) These tasks continue to remain a challenge in understanding the structural interplay between the different elements of a chart. Effective visual language understanding in particular requires two key processes: (a) comprehensive semantic layout understanding of the chart; (b) robust statistical reasoning to accurately capture and analyze the underlying data \cite{liu2023matcha}.

In light of these challenges, we investigate zero-shot and training-free approaches for VLMs in chart summarization.
%
%
Program-of-Thoughts (PoT) \cite{chen2023programofthought} is a zero-shot prompting method, which was originally proposed to disentangle computation from reasoning to augment a model’s statistical reasoning capability.
The success of PoT in chart question answering (QA) \cite{zhang2024tinychart} with Python programs has motivated our exploration of chart summarization, investigating the effectiveness of the PoT guiding VLMs to perform numerical computations and logical reasoning via Python programs as intermediate steps in the chart summarization process, which focuses on generating more structurally complex and extensive sentences, rather than just concise answers.
%
%
%
Instead of relying on the provided real chart data tables for PoT in recent PoT research works, we acknowledge that in real-world scenarios, most charts lack accompanying data tables. Therefore, we investigate a  PoT strategy pipeline for chart summarization with simultaneously generated chart data tables. Our key contributions are as follows:

\begin{itemize}
    \item We propose a PoT-integrated, training-free pipeline, enhancing lightweight VLMs for chart summarization in a zero-shot learning setting.
    \item We demonstrate the PoT prompting strategy outperforms Direct and MCoT approaches in certain scenarios, particularly across diverse types of VLMs, charts, and supplementary textual data in chart summarization.
    \item We conduct comprehensive evaluations across lexical, semantic, and factual dimensions to validate the effectiveness of the PoT prompting strategy for chart summarization.
\end{itemize}

%


%

%

\section{Literature Review}


\subsection{Chart Understanding}
\noindent\textbf{Template-Based}
%
Early approaches to automatic chart understanding, particularly the sub-task of chart summarization, often relied on planning-based architecture and template-based generation methods \cite{mittal1998templatebased, fasciano2000templatebased, green2004templatebased, reiter2007templatebased, ferres2007improving, ferres2013templatebased}.
%
Recent template-based research has focused on utilizing statistics (e.g., min, max, trends) from chart numerical data for presenting the facts \cite{demir2012templatestatis, cui2019templatestatis, srinivasan2019templatestatis, wang2020templatestatis}, forming the statistics analysis into textual summarization output.
Some research utilized the off-the-shelf OCR (Optical Character Recognition) tools or detectors to represent chart data into textual tables and other representations, relying on pipeline methods \cite{singh2020ocrc2t, sidorov2020ocrc2t, methani2020ocrc2t, hu2021ocrc2t, fu2022ocrc2t,kantharaj2022chart,liu2023ocrc2t}.
More recently, ResNet \cite{he2016resnet} encoder and LSTM decoder were used to process the chart and create the caption \cite{chen2020templatelstm}.
However, compared to data-driven models, template-based approaches struggle with complex visual patterns and numerical reasoning, with high costs in producing generics and matching variations in vocabulary choices.

\noindent\textbf{Pretrained}
With the progression of deep learning techniques, which subsequently improved general computer vision using neural networks and Transformer \cite{vaswani2017attention}, recent work began to adopt encoder-decoder architectures to improve chart understanding \cite{wang2025multimodalmcot}, including Transformer \cite{singh2020transformerc2t, obeid2020chart2text, kantharaj2022chart, lee2023pix2struct}, LSTM \cite{spreafico2020lstmc2t}, CNN+LSTM \cite{hsu2021cnnlstmc2t}, and VLMs \cite{liu2023matcha},
%
%
%
which are pre-trained on both visual and text data, often with specialized text and image encoders, and have shown significant promise in tasks requiring joint understanding of multiple modalities.
However, challenges remain in grounding the factual and logical coherence in generated summaries, particularly when dealing with complex charts requiring numerical reasoning.

\begin{figure*}[ht]
    \centering
    \includegraphics[width=\linewidth]{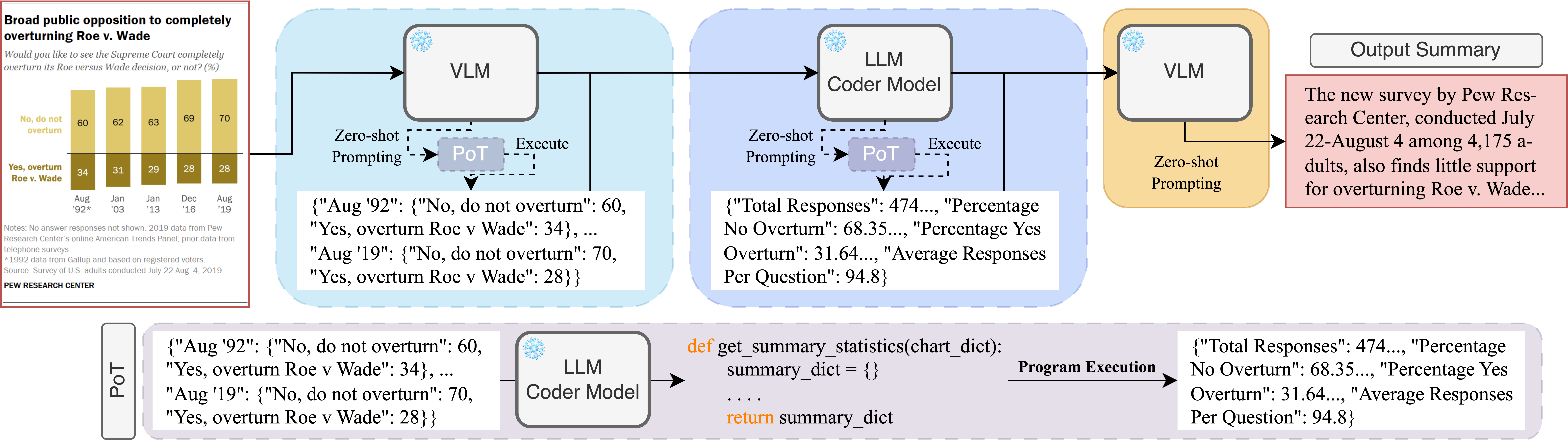}
    \caption{Process of implementing the Program of Thought (PoT) given a chart. It can be seen as a process of enhancing statistical reasoning to extract summary statistics, typically total counts, minimum, and maximum values from the chart, along with labels that contain the numerical values.}
    \label{fig:c2t_pipeline}
\end{figure*}


\noindent\textbf{Fine-Tuned}
Aside from pre-training the model, fine-tuning the pre-training model \cite{tang2023vistext} and instruction fine-tuning \cite{ouyang2022finetuning} have also become widely adopted as an alternative to improve the performance of LLMs and VLMs \cite{liu2023matcha, zhou2023enhancedchartT5, masry2023unichart, han2023chartllama, ko2024chartllm, huang2024lvlms, masry2024chartinstruct, meng2024chartassistant, zhang2024tinychart, liu2024mmc, liu2024chartthinker,masry2025chartgemma}. Instruction tuning is used to generalize the language capability of the model, reducing repetitions and hallucinations generated in summarization compared to pre-training approaches \cite{meng2024chartassistant}.
However, these methods typically rely on the data tables of charts, failing to capture the nuance of the visual artifacts present in charts. Furthermore, their heavy parameter sizes present notable challenges for deployment in computationally constrained environments.

\subsection{Chart Representations}
Representing the chart in structured data, the chart-to-table \cite{meng2024chartassistant} task represents it in the tabular format, 
but often comes at the cost of losing finer details in the chart.
%
Performing similarly to data tables, scene graphs are easily formatted for web-based charts \cite{tang2023vistext}.
%
Code format is considered, and existing methodologies define two typical chart-to-code approaches: (1) Chart Derendering \cite{liu2023matcha, lee2023pix2struct}; and (2) Program of Thoughts \cite{chen2023programofthought, zhang2024tinychart}. However, codes mainly aim to run for the chart recreation or question answering tasks on narrowly defined questions, rather than representing the whole chart. 
This paper proposes an auxiliary task of chart-to-table, which is chart-to-dictionary in Python code format, which uses VLM's chart understanding capability to represent the chart as a Python dictionary.

\subsection{Prompting}
%
Inspired by the success of Chain-of-Thought (CoT) prompting \cite{wei2022cot} for improving reasoning capabilities, researchers are extending similar mechanisms to VLMs for chart understanding, seeking to mirror the human cognitive process of visual analysis.
%
%
This is achieved through multimodal-purpose prompting Multimodal Chain of Thought (MCoT) \cite{wang2025multimodalmcot,liu2024chartthinker} reasoning, which extends the rationale from texts to visual modalities \cite{choi2025end}.
%
%
%
%
To contrast with MCoT, PoT \cite{chen2023programofthought,luo2024pythonpot} intermediate reasoning steps are articulated as executable programs, while executing the program to generate reasoning and statistical computation about the chart data in complex numerical reasoning tasks.

In this work, our pipeline method builds upon these advancements by focusing on PoT prompting in zero-shot chart summarization. By extending the PoT concept to the visual domain of charts, it could decrease hallucinations that language models typically have when outputting calculations, as it provides more explicit and verifiable numeric reasoning processes for VLMs \cite{zhang2024tinychart}, potentially leading to more accurate and factually grounded summaries by delegating complex calculations to a code interpreter.
This work differentiates itself from existing works by specifically investigating the benefits and limitations of generating executable code as intermediate reasoning steps for chart summarization with lightweight VLMs.



\section{Method}


We propose a pipeline with the PoT integrated to augment a VLM's capability for statistical reasoning on chart data summarization. An illustration of the proposed PoT-integrated chart summarization pipeline is presented in Figure~\ref{fig:c2t_pipeline}.
Our prompts can be found in Appendix~\ref{sec:apix_prompts}.


\begin{figure}[t]
    \centering
    \includegraphics[width=\linewidth]{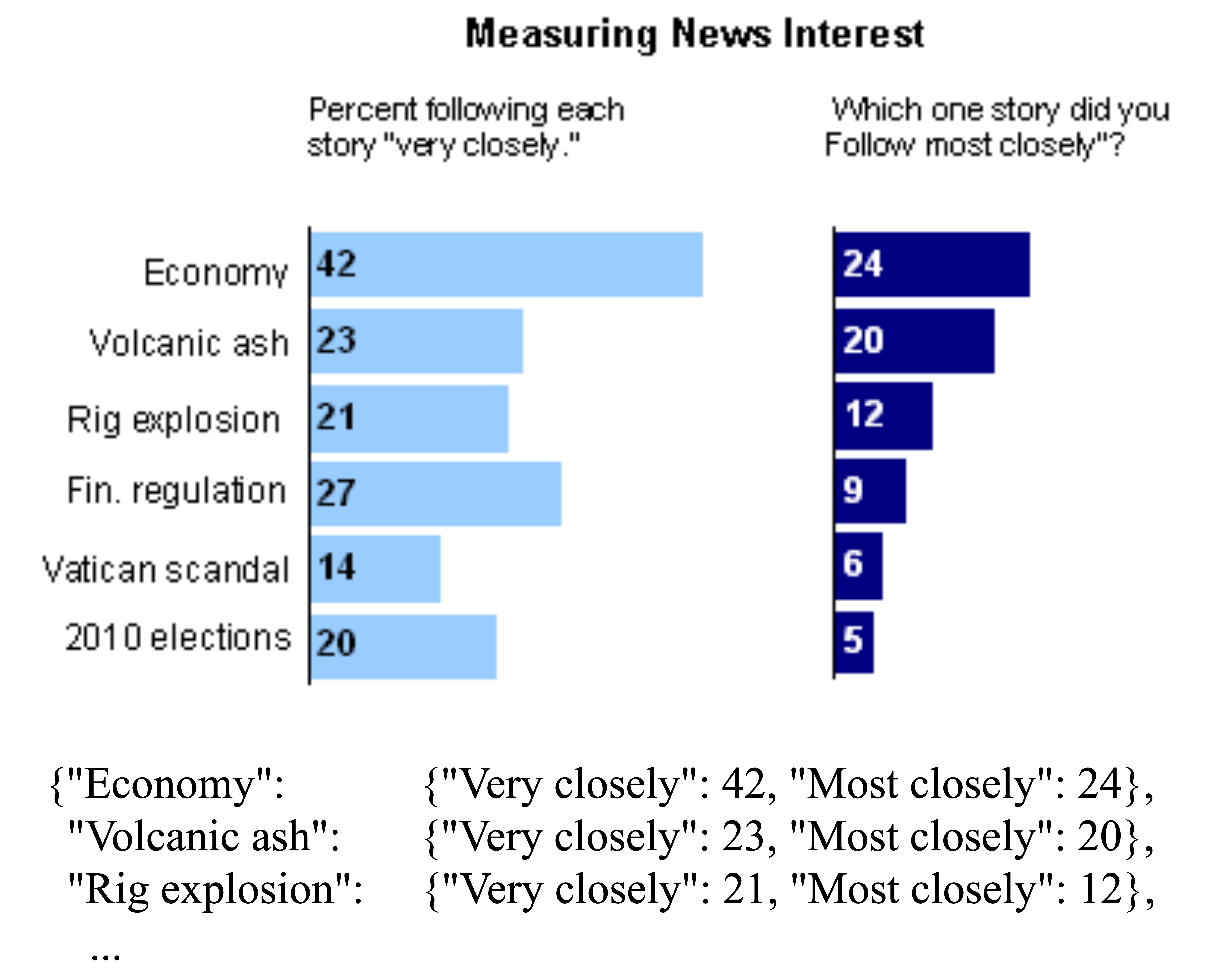}
    \caption{Representing chart (top) as a Python dictionary (bottom).}
    \label{fig:chart_dict}
\end{figure}

\subsection{Chart Representation as a VLM-Generated Python Dictionary}
In order for the chart to interface with the code, the chart needs to be represented in a manner that can interact with the Python interpreter. As shown in Figure~\ref{fig:chart_dict}, Python dictionaries can represent the code in a more free-form structure, allowing for grounding the values compared to the data table, which is more flexible compared to a markdown table and usable by the LLM-generated program.
%
However, lightweight VLMs can struggle to create executable Python code, which consists of wrong syntax, incomplete messages, and even meaningless code-agnostic terminologies when facing the complex code generation request, adding noise. Given that, aside from reflecting understanding from charts, the code needs to be valid and executable.
In Appendix \ref{sec:apix_failure_case_study}, we list more details of the failure case analysis.
To handle failure cases in dictionary generation, we mainly used InternVL-2.5-4B \cite{chen2024internvl} on dictionary generation in a zero-shot setting, and if the generated Python dictionary is not executable, it is converted with ChatGPT (GPT-4o-mini) \cite{openai2024gpt4omini} instead.

\subsection{Statistical Analysis with PoT Prompting}
Since the chart is represented as a Python dictionary, it can be more free-form in containing data and being passed to a Python program.
We adopt a similar methodology described in PoT, as the work \cite{chen2023programofthought} stated that the program’s line-by-line structure acts as a proxy for the numerical reasoning steps of the model.
Code is passed to an LLM to generate a program to do statistical analysis as an intermediate result to provide more context for chart summarization. Compared to QA as a task, statistical analysis with PoT demonstrates numerical reasoning since it demonstrates how the models understand which data points or statistics are necessary to create summary statistics. 
This paper uses Qwen-2.5-Coder-14B \cite{hui2024qwen25coder} for the complex statistics code generation conversion. The LLM is instructed to generate a Python program using the Python dictionary in the prompt to generate summary statistics relevant to the chart dictionary. This adapts PoT for the chart summarization task as the generated program provides more context to be used for text generation while providing accurate calculations. Code generated by the LLM is constrained to use only the functions from Python’s built-in library. To validate and execute the generated Python program by the PoT strategy, we used the built-in \codeword{exec} function in Python for automatic code validation.


\subsection{Program Execution}
The generated Python program for statistics calculation is executed using a Python interpreter. This step ensures the accuracy of the statistical results, mitigating potential errors that LLMs might make when generating tokens through direct calculations. The program returns a Python statistics dictionary that contains key-value pairs of the summary statistics and the calculated values.
At the end, the statistical results in our pipeline are input with the chart into a VLM to assist the chart summarization task.

\section{Experiment}

We present our experimental setup in Appendix~\ref{sec:apix_experiment}. The overview of our datasets, evaluation metrics, baseline methods, and benchmark and backbone models is provided in the following subsections.



\noindent \textbf{Evaluation.}
We evaluated PoT prompting for chart summarization on both the test sets of the Pew \cite{kantharaj2022chart} and VisText \cite{tang2023vistext}, following the previous evaluation works \cite{masry2023unichart, meng2024chartassistant} for evaluating the PoT on varying degrees of complex charts to show its generalizability. 
The VisText is built upon Statista \cite{kantharaj2022chart} with richly labelled L2/L3 captions.
Chart type distributions of datasets are summarized in Table~\ref{tab:pew_vistext}, which across a variety of simple and complex charts. More details on the dataset statistics and topic distribution information are presented in the Appendix~\ref{sec:apix_dataset}.
To evaluate the effectiveness of the methods, we employ BLEU \cite{post2018callbleu} and CIDEr \cite{vedantam2015cider} as the evaluation metric following previous works \cite{kantharaj2022chart,liu2023matcha,masry2023unichart,meng2024chartassistant}.
Additionally, we use F1 scores of ROUGE \cite{lin2004rouge} and BERTScore \cite{zhang2020bertscore} for semantic evaluation; UniEval \cite{zhong2022unieval} and AlignScore-large \cite{zha2023alignscore} for factual evaluation; and human evaluation with 3 human evaluators. We provide details of evaluation metrics in the Appendix~\ref{sec:apix_evaluation} and Appendix~\ref{sec:apix_human_evaluation}.

\begin{table}[h]
\centering
\resizebox{\linewidth}{!}{%
\begin{tabular}{lcccccc}
\toprule
\multirow{2}{*}{\textbf{Type}}  & \multicolumn{3}{c}{\textbf{Pew}} & \multicolumn{3}{c}{\textbf{VisText}} \\
\cmidrule(lr){2-7}
      & Simp. & Comp. & All & Simp. & Comp. & All \\
\midrule
Area       & 7      & 13   & 20   & 157  & 81    & 238 \\
Bar        & 128    & 840  & 968  & 303  & 128   & 431 \\
Line       & 37     & 312  & 349  & 135  & 78    & 213 \\
Pie        & 41     & 0    & 41   & 0    &  0   & 0 \\
Scatter    & 0      & 15   & 15   & 0    &  0   & 0 \\
\midrule
Total      & 213  & 1,180  & 1,393  & 595 & 287  & 882 \\
\bottomrule
\end{tabular}
}
\caption{Distribution of chart types by \textbf{Simp}le and \textbf{Comp}lex complexities of the Pew and VisText datasets.}
\label{tab:pew_vistext}
\end{table}

\noindent \textbf{Baselines.}
We compared two other types of prompting strategies as baselines: (1) Directly prompting (Direct) the model to summarize the chart, given that this approach is also what is done by fine-tuned end-to-end models \cite{huang2024pixels, liu2024chartthinker}; (2) Multimodal CoT (MCoT), which adheres to the framework in \cite{wang2025multimodalmcot}, prompting to return an outline of all key information and trends derived from the chart.



\noindent \textbf{Backbones.}
To understand the effects of the PoT, we compared (1) Existing models and methods in the chart-to-text domain: (a) Pretrained Chart-To-Text Models: OCR-Field-Infuse \cite{chen2020fieldinfusing, kantharaj2022chart}, Monkey \cite{li2023monkey}; (b) Prefix-tuning Chart-To-Text Models: image-scene-graph-PT \cite{tang2023vistext}, image-data-table-PT \cite{tang2023vistext}; (c) Commonly used VLMs: Blip2-flant5xl \cite{li2023blip2}, Qwen-VL \cite{bai2023Qwen-VL}; and (2) Lightweight VLMs: DeepSeek (DeepSeek-VL2-tiny) \cite{wu2024deepseekvl2tiny}, InternVL (InternVL-2.5-4B) \cite{chen2024internvl}, LLaVA (LLaVA-v1.6-mistral-7B-hf) \cite{liu2023llavav16mistral7bhf}, and Qwen (Qwen2.5-VL-3B-Instruct) \cite{qwen2025qwen25vl} on the representative datasets of Pew and VisText. All experiments were done with the zero-shot setting models.

\begin{table}[t]
  \centering
  \resizebox{\linewidth}{!}{%
    \begin{tabular}{lccc*{8}{c}}
      \toprule
      \multirow{2}{*}{\textbf{Method}}
        & \multicolumn{2}{c}{\textbf{Pew}}
        & \multicolumn{2}{c}{\textbf{VisText}} \\
    \cmidrule(lr){2-5}
      &
        BLEU & CIDEr & BLEU & CIDEr \\
      \midrule
    OCR-Field-Infuse        & 0.2 & 0.3 & 0.3 & - \\
    Monkey                  & 0.4 & 1.7 & - & - \\
    Qwen-VL-9.6B            & 0.5 & 2.6 & - & - \\
    Blip2-flant5xl-4B       & 0.2 & 0.8 & - & - \\
    image-scene-graph-PT    & - & - & 0.3 & - \\
    image-data-table-PT     & - & - & 0.3 & - \\
    \midrule
    Qwen2.5-VL-3B+PoT       & \textbf{3.1} & 0.1 & \textbf{1.7} & 0.1 \\
      \bottomrule
    \end{tabular}%
    }
    \caption{We compare our PoT-adopted zero-shot VLM (Qwen2.5-VL-3B+PoT) with different chart summarization methods on Pew and VisText test datasets. We referenced the results from Chart-To-Text \cite{kantharaj2022chart}, VisText \cite{tang2023vistext}, and ChartAssistant \cite{meng2024chartassistant}.}
  \label{tab:comparison_results}
\end{table}



\begin{table*}[t]
  \centering
  \renewcommand{\arraystretch}{1.2}
  \resizebox{\textwidth}{!}{%
    \begin{tabular}{ll*{19}{r}}
      \toprule
         \multirow{3}{*}{\shortstack{\textbf{VLM} \\  \\ \textbf{-Prompting}}}
        & \multicolumn{12}{c}{\textbf{Pew}}
        & \multicolumn{8}{c}{\textbf{VisText}} \\
      \cmidrule(lr){2-13}\cmidrule(lr){14-21}
      &
          \multicolumn{2}{c}{Area}
        & \multicolumn{2}{c}{Bar}
        & \multicolumn{2}{c}{Line}
        & \multicolumn{2}{c}{Pie}
        & \multicolumn{2}{c}{Scatter}
        & \multicolumn{2}{c}{All}
        & \multicolumn{2}{c}{Area}
        & \multicolumn{2}{c}{Bar}
        & \multicolumn{2}{c}{Line}
        & \multicolumn{2}{c}{All} \\
      &
          BLEU & CIDEr
        & BLEU & CIDEr
        & BLEU & CIDEr
        & BLEU & CIDEr
        & BLEU & CIDEr
        & BLEU & CIDEr
        & BLEU & CIDEr
        & BLEU & CIDEr
        & BLEU & CIDEr
        & BLEU & CIDEr \\
      \midrule
      \textbf{deepseek-vl2-tiny} \\

   ZeroShot-Direct 
  & 1.9682 & 0.0427 & 2.6653 & 0.0608 & 1.7169 & 0.0471 & 4.5805 & 0.1391 
  & 0.7646 & 0.0412 & 2.4676 & 0.0591 
  & 1.8347 & 0.0920 & 1.5262 & 0.0731 & 2.0429 & 0.0851 & 1.7346 & 0.0824 \\

   ZeroShot-MCoT  
  & 1.6352 & 0.0526 & 1.8918 & 0.0403 & 1.2924 & 0.0360 & 3.1608 & 0.0671 
  & 1.0925 & 0.0657 & 1.7658 & 0.0399 
  & 0.9308 & 0.0410 & 0.7613 & 0.0353 & 1.1508 & 0.0388 & 0.9001 & 0.0380 \\


   ZeroShot-PoT  
  & 0.1254 & 0.0018 & 0.2767 & 0.0127 & 0.2736 & 0.0173 & 0.2496 & 0.0190 
  & 0.2219 & 0.0005 & 0.2746 & 0.0135 
  & 0.8102 & 0.0710 & 0.3489 & 0.0523 & 0.5821 & 0.0685 & 0.5603 & 0.0615 \\
\midrule
\textbf{internVL-2.5} \\

         ZeroShot-Direct 
  & 3.6507 & 0.0426 & 3.5832 & 0.0318 & 2.7521 & 0.0296 & 4.6431 & 0.1025 
  & 2.6224 & 0.0001 & 3.4041 & 0.0328 
  & 1.1306 & 0.0125 & 0.9387 & 0.0088 & 1.3401 & 0.0212 & 1.0808 & 0.0130 \\

        ZeroShot-MCoT  
  & 2.3817 & 0.0257 & 2.0626 & 0.0106 & 1.4369 & 0.0061 & 1.9856 & 0.0053 
  & 1.5318 & 0.0003 & 1.9113 & 0.0094 
  & 0.8414 & 0.0022 & 0.8978 & 0.0005 & 1.0359 & 0.0030 & 0.9175 & 0.0015 \\
  
         ZeroShot-PoT  
  & 2.8535 & \textbf{0.0713} & 1.9995 & \textbf{0.0664} & 1.9136 & \textbf{0.0404} & 2.0840 & 0.0907 
  & 1.3768 & \textbf{0.0819} & 1.9896 & \textbf{0.0603} 
  & 1.1281 & \textbf{0.0246} & 0.9299 & \textbf{0.0172} & \textbf{1.6892} & \textbf{0.0274} & \textbf{1.1736} & \textbf{0.0212} \\
\midrule
\textbf{llava-NeXT}  \\

         ZeroShot-Direct 
  & 4.8807 & 0.1561 & 5.7756 & 0.1069 & 4.6735 & 0.1133 & 7.8216 & 0.2135 
  & 4.3993 & 0.0074 & 5.5350 & 0.1107 
  & 2.6597 & 0.0272 & 2.5564 & 0.0334 & 3.4469 & 0.0612 & 2.7918 & 0.0384 \\

          ZeroShot-MCoT  
  & 6.1606 & 0.0329 & 5.9175 & 0.0928 & 4.6181 & 0.0644 & 5.7460 & 0.1498 
  & 3.9118 & 0.0808 & 5.6347 & 0.0869 
  & 2.5957 & 0.0478 & 2.2776 & 0.0243 & 3.5833 & 0.0499 & 2.6622 & 0.0365 \\

          ZeroShot-PoT  
  & 3.1421 & 0.1069 & 4.1897 & 0.1027 & 3.5534 & 0.0925 & 2.7975 & 0.0895 
  & 3.3424 & \textbf{0.1210} & 3.9888 & 0.0996 
  & 2.3603 & 0.0321 & 2.2635 & \textbf{0.0457} & 2.9584 & 0.0580 & 2.4604 & \textbf{0.0448} \\
\midrule
\textbf{qwen2.5-VL-3B}  \\

       ZeroShot-Direct 
  & 1.9350 & 0.0523 & 3.6251 & 0.1002 & 2.5562 & 0.0643 & 5.9420 & 0.1384 
  & 2.0714 & 0.0272 & 3.3929 & 0.0905 
  & 2.6399 & 0.1481 & 2.1772 & 0.0979 & 3.1147 & 0.1519 & 2.4984 & 0.1254 \\

       ZeroShot-MCoT  
  & 1.4980 & 0.0735 & 2.6168 & 0.0814 & 1.8583 & 0.0602 & 3.7722 & 0.2156 
  & 1.5976 & 0.0431 & 2.4388 & 0.0794 
  & 1.5847 & 0.0837 & 1.3648 & 0.0791 & 1.9742 & 0.0707 & 1.5783 & 0.0782 \\
  
       ZeroShot-PoT  
  & \textbf{3.3383} & 0.0409 & 3.3091 & 0.0734 & 2.3678 & 0.0597 & 3.8250 & 0.1662 
  & 1.0761 & 0.0203 & 3.0906 & 0.0712 
  & 1.6593 & 0.0780 & 1.4806 & 0.0801 & 2.0928 & 0.0890 & 1.6639 & 0.0826 \\

      \bottomrule
    \end{tabular}%
  }
  \caption{Evaluation results of VLMs on different prompting methods on Pew and VisText datasets evaluated on BLEU and CIDEr scores.}
  \label{tab:ablation}
\end{table*}


\section{Results \& Discussion}

\subsection{PoT Approach Against Existing Chart-To-Text Models}

Table~\ref{tab:comparison_results} shows the BLEU and CIDEr scores for each model on the Pew and VisText datasets. We referenced evaluation results from Chart-To-Text \cite{kantharaj2022chart}, VisText \cite{tang2023vistext}, and ChartAssistant \cite{meng2024chartassistant}.
As shown in the table, we observed that our PoT prompting approach overperforms baseline Chart-To-Text methods in the BLEU evaluation scores, but underperforms in the CIDEr evaluation scores. This may be due to CIDEr placing more emphasis on important and rare words, as it calculates TF-IDF weighted n-gram similarity. While BLEU also focuses only on surface-level word matching and ignores semantic consistency, we subsequently evaluate our PoT prompting approach using BERTScore and ROUGE to capture semantic relevance, and UniEval and AlignScore to assess factual correctness beyond lexical overlap.

\subsection{PoT Approach Against Baselines}

We evaluate baseline prompting strategies and our PoT prompting strategy and report results of our experiment in Table~\ref{tab:ablation} and Table~\ref{tab:ablation_asueval}. Full experimental results and extended ablation studies are in the Appendix~\ref{sec:ablation_study_tests}. Across the evaluated models, the impact of the PoT prompting strategy varied significantly with models and chart types. We observed instances where the PoT led to substantial improvements in performance, while in other cases, its impact was less pronounced or even negative compared to the Direct and MCoT approaches.

\noindent\textbf{PoT Effectiveness Against Chart Types.}
We notice that the results from different charts are varied, and we suppose this may be due to the uniqueness of each chart structure, texts included in the chart, chart data size, and data complexity. For example, in the case of the Qwen2.5-VL model, the BLEU score increases from 1.94 to 3.34 with PoT, demonstrating the effectiveness of the PoT strategy in enhancing information collection from area charts, which are with limited data information.

\begin{table*}[t]
  \centering
  \renewcommand{\arraystretch}{1.2}
  \resizebox{\textwidth}{!}{%
    \begin{tabular}{ll*{19}{r}}
      \toprule
         \multirow{3}{*}{\shortstack{\textbf{VLM} \\  \\ \textbf{-Prompting}}}
        & \multicolumn{12}{c}{\textbf{Pew}}
        & \multicolumn{8}{c}{\textbf{VisText}} \\
      \cmidrule(lr){2-13}\cmidrule(lr){14-21}
      &
          \multicolumn{2}{c}{Area}
        & \multicolumn{2}{c}{Bar}
        & \multicolumn{2}{c}{Line}
        & \multicolumn{2}{c}{Pie}
        & \multicolumn{2}{c}{Scatter}
        & \multicolumn{2}{c}{All}
        & \multicolumn{2}{c}{Area}
        & \multicolumn{2}{c}{Bar}
        & \multicolumn{2}{c}{Line}
        & \multicolumn{2}{c}{All} \\
      &
          AS-l & UE-o
        & AS-l & UE-o
        & AS-l & UE-o
        & AS-l & UE-o
        & AS-l & UE-o
        & AS-l & UE-o
        & AS-l & UE-o
        & AS-l & UE-o
        & AS-l & UE-o
        & AS-l & UE-o \\
      \midrule
      \textbf{deepseek-vl2-tiny} \\

   ZeroShot-Direct 
  & 13.17 & \textbf{79.00} & \textbf{26.04} & \textbf{76.55} & \textbf{16.35} & \textbf{78.61} & \textbf{27.69} & \textbf{81.54} 
  & \textbf{27.80} & \textbf{83.14} & \textbf{23.50} & \textbf{77.32} 
  & \textbf{7.27} & \textbf{84.34} & \textbf{5.30} & 78.45 & \textbf{7.78} & \textbf{84.36} & \textbf{6.43} & \textbf{81.47} \\

   ZeroShot-MCoT  
  & 11.36 & 74.54 & 19.12 & 75.62 & 13.26 & 74.45 & 17.80 & 79.50 
  & 20.68 & 80.46 & 17.52 & 75.48 
  & 4.22 & 80.91 & 3.67 & \textbf{79.81} & 6.14 & 80.90 & 4.41 & 80.37 \\


   ZeroShot-PoT  
  & \textbf{15.80} & 51.09 & 16.00 & 55.53 & 14.61 & 52.43 & 13.97 & 56.76 
  & 5.63 & 53.43 & 15.47 & 54.70
  & 3.76 & 58.44 & 3.30 & 57.04 & 2.85 & 56.02 & 3.31 & 57.17 \\
\midrule
\textbf{internVL-2.5} \\

         ZeroShot-Direct 
  & 12.02 & 75.29 & 25.30 & 77.44 & 19.36 & 78.32 & 27.34 & \textbf{81.97} 
  & 13.18 & 82.42 & 23.55 & 77.82 
  & 6.15 & 84.00 & 5.71 & 82.39 & 8.52 & 83.62 & 6.51 & 83.12 \\

        ZeroShot-MCoT  
  & 10.29 & \textbf{79.57} & 18.75 & 76.37 & 12.97 & 76.10 & 18.70 & 77.04 
  & 14.59 & 74.93 & 17.13 & 76.35 
  & 4.32 & 81.67 & 3.92 & 81.41 & 4.73 & 81.38 & 4.22 & 81.47 \\
  
         ZeroShot-PoT  
  & \textbf{25.91} & 78.48 & \textbf{37.60} & \textbf{84.26} & \textbf{36.74} & \textbf{83.73} & \textbf{27.96} & 81.63 
  & \textbf{31.95} & \textbf{87.96} & \textbf{36.87} & \textbf{84.01}
  & \textbf{10.79} & \textbf{86.60} & \textbf{7.71} & \textbf{86.55} & \textbf{10.76} & \textbf{86.06} & \textbf{9.28} & \textbf{86.44} \\
\midrule
\textbf{llava-NeXT}  \\

         ZeroShot-Direct 
  & \textbf{19.51} & \textbf{84.26} & 28.07 & \textbf{83.43} & 22.91 & \textbf{84.76} & \textbf{27.84} & 82.69 
  & 22.38 & 81.22 & 26.59 & \textbf{83.73} 
  & \textbf{10.86} & \textbf{87.53} & \textbf{6.00} & \textbf{86.84} & \textbf{7.07} & \textbf{87.73} & \textbf{7.47} & \textbf{87.24} \\

          ZeroShot-MCoT  
  & 11.66 & 83.94 & $/$ & $/$ & 20.00 & 84.31 & 21.45 & \textbf{85.70} 
  & 17.65 & \textbf{87.04} & $/$ & $/$ 
  & $/$ & $/$ & 4.99 & 86.29 & $/$ & $/$ & $/$ & $/$ \\

          ZeroShot-PoT  
  & 16.90 & 71.24 & \textbf{31.14} & 83.04 & \textbf{28.15} & 81.77 & 27.67 & 82.10 
  & \textbf{25.21} & 83.99 & \textbf{30.02} & 82.54 
  & 5.18 & 86.21 & 5.38 & 85.53 & 6.73 & 85.92 & 5.65 & 85.81 \\
\midrule
\textbf{qwen2.5-VL-3B}  \\

       ZeroShot-Direct 
  & 19.93 & 80.57 & 35.17 & \textbf{84.30} & 23.08 & \textbf{81.29} & 37.93 & 87.19 
  & 23.09 & \textbf{90.38} & 31.87 & \textbf{83.64} 
  & 7.74 & 84.92 & 6.75 & 82.17 & 10.64 & 84.38 & 7.96 & 83.45 \\

       ZeroShot-MCoT  
  & 17.17 & 78.40 & \textbf{36.34} & 82.75 & 24.14 & 80.99 & \textbf{46.65} & \textbf{87.55} 
  & 21.83 & 90.12 & \textbf{33.16} & 82.47 
  & 9.76 & \textbf{85.62} & \textbf{7.72} & \textbf{82.61} & 12.58 & \textbf{85.90} & 9.44 & \textbf{84.22} \\
  
       ZeroShot-PoT  
  & \textbf{26.79} & \textbf{81.39} & 32.09 & 79.64 & \textbf{26.93} & 78.87 & 40.71 & 85.56 
  & \textbf{24.60} & 86.61 & 30.89 & 79.72 
  & \textbf{10.99} & 82.32 & 6.60 & 80.58 & \textbf{13.65} & 82.60 & \textbf{9.49} & 81.54 \\

      \bottomrule
    \end{tabular}%
  }
  \caption{Evaluation results of VLMs on different prompting methods on Pew and VisText datasets evaluated on
\textbf{A}lign\textbf{S}core-\textbf{l}arge and \textbf{U}ni\textbf{E}val-\textbf{o}verall scores.}
  \label{tab:ablation_asueval}
\end{table*}

\noindent\textbf{PoT Effectiveness Against VLMs.}
Regarding influences by VLMs, for the DeepSeek-vl2-tiny model, the application of the PoT resulted in considerably lower scores across all reported metrics compared to both the Direct and MCoT methods. This suggests that for this particular model architecture, the PoT strategy in its current implementation might not be beneficial or could even hinder performance on the evaluated tasks. This reveals that the PoT strategy may introduce additional noise or mislead the emphasized information, and may interfere with the model's original processing and understanding of the chart.
In contrast, the InternVL-2.5 model demonstrated a more nuanced response to the PoT prompting strategy. While the Direct method often yielded the highest scores, the PoT strategy achieved comparable or even slightly better results on certain metrics compared to the MCoT strategy in most cases. For example, the PoT strategy achieved a BLEU score of 2.85, which is lower than the Direct method (3.65) but higher than the MCoT strategy (2.38) of the area charts in the Pew dataset. Even on considering all chart types, these trends hold. This indicates that for InternVL-2.5, the PoT strategy can be a viable alternative to the MCoT strategy in certain scenarios, especially from the results of factual evaluation. Similarly, LLaVA-NeXT also had a mixed response given the two datasets, where no conclusive trends can be observed between the different prompting methods. One interesting observation from this comparison is that while the BLEU values of the PoT strategy are lower than the other methods, on average, it outperforms the other prompting techniques on CIDEr, AlignScore or UniEval, indicating some of its effectiveness in these cases.
We suggest that this may arise from the inherent design and pertaining data differences in the VLMs with respect to chart understanding. Specifically, DeepSeek-vl2 is equipped with a dedicated vision encoder and a vision-language adapter, originally designed to optimize performance on visual tasks such as chart interpretation.
In contrast, InternVL-2.5 is built upon a Vision Transformer architecture integrated with a large language model, while pre-trained with one of the benchmark datasets \cite{chen2024internvl}, the VisText dataset, placing more confidence on the fusion of textual information in the chart-to-text task. As a result, when we enlarge the textual information using the PoT strategy, the performance outcomes of DeepSeek-vl2 and InternVL-2.5 can diverge, potentially yielding opposite trends. This observation suggests that the PoT strategy does not universally benefit all VLMs in chart summarization, but is particularly advantageous for those that emphasize textual information.

\noindent\textbf{PoT Compared with MCoT.}
On the other hand, Qwen-2.5VL-3B showed that the PoT strategy consistently outperformed the MCoT strategy while underperforming relative to the Direct prompting. This suggests that for the Qwen2.5-VL-3B model, the PoT strategy appears to be a more effective CoT prompting strategy compared to the standard MCoT approach across the evaluated tasks.
This may be due to the PoT strategy introducing more new statistical content into the chart summarization process during chart data interpretation compared to the MCoT approach. While the PoT generates additional statistical information, MCoT primarily offers a high-level data outline and trends.




\begin{table}[h]
\centering
\resizebox{\linewidth}{!}{%
\begin{tabular}{lccc}
\toprule
 & DeepSeek & InternVL & Qwen  \\
\midrule
Template-based & 18.67  & 19.00   & 22.67 \\
PoT-based      & \textbf{31.33} & \textbf{31.00}  & \textbf{27.33} \\
\bottomrule
\end{tabular}
}
\caption{Human evaluation.}
\label{tab:huamn_eval}
\end{table}




\subsection{PoT Approach Against VLM Backbones}

While the PoT strategy demonstrated potential for improving performance, particularly for the InternVL-2.5 and Qwen2.5-VL-3B models in certain scenarios, we conducted further investigations to valid the effectiveness of adopting PoT and identify the potential information factors contributing to the varying effectiveness of the PoT strategy and to estimate the extent to which information influences the performance of the PoT strategy pipeline.
We compared: (1) Template-based: using the predefined Python program template; and (2) PoT-based: using the PoT for generating the statistics dictionary in chart summarization with human evaluation, as shown in Table~\ref{tab:huamn_eval}.
In addition, we conducted a series of experiments focusing on the textual components that serve as supplementary inputs to the VLM alongside the input chart. The experimental settings are as follows: (1) Title: Use only the title as input to the VLM, without applying the PoT strategy; (2) Dict+Title: Use the PoT-generated Python dictionary along with the title as input to the VLM; (3) Stats+Title: Use the PoT strategy to generate a statistics dictionary, combined with the title as input to the VLM; (4) Dict+Stats+Title: Use the full set of inputs, including the PoT-generated Python dictionary, the PoT-generated statistics dictionary, and the title as input to the VLM; (5) Dict+StatsT+Title: Replace the LM with a predefined Python program template for generating the statistics dictionary, and use the generated statistics dictionary together with the Python dictionary and title as input to the VLM. The experimental results that were evaluated on ROUGE-L and BERTScores are illustrated in Table~\ref{tab:ablation_rouge_bert}.

\begin{table*}[t]
  \centering
  \renewcommand{\arraystretch}{1.2}
  \resizebox{\textwidth}{!}{%
    \begin{tabular}{ll*{20}{r}}
      \toprule
       \multirow{3}{*}{\shortstack{\textbf{VLM} \\  \\ \textbf{+Textual Data}}}
        & \multicolumn{12}{c}{\textbf{Pew}}
        & \multicolumn{8}{c}{\textbf{VisText}} \\
      \cmidrule(lr){2-13}\cmidrule(lr){14-21}
      &
         \multicolumn{2}{c}{Area}
        & \multicolumn{2}{c}{Bar}
        & \multicolumn{2}{c}{Line}
        & \multicolumn{2}{c}{Pie}
        & \multicolumn{2}{c}{Scatter}
        & \multicolumn{2}{c}{All}
        & \multicolumn{2}{c}{Area}
        & \multicolumn{2}{c}{Bar}
        & \multicolumn{2}{c}{Line}
        & \multicolumn{2}{c}{All} \\
      &
         R-L & BS
        & R-L & BS
        & R-L & BS
        & R-L & BS
        & R-L & BS
        & R-L & BS
        & R-L & BS
        & R-L & BS
        & R-L & BS
        & R-L & BS \\
      \midrule
      \textbf{deepseek-vl2-tiny}  \\

 Title
  & \textbf{13.57} & \textbf{84.78} & \textbf{13.57} & \textbf{85.49} & \textbf{12.26} & \textbf{84.83} & \textbf{16.98} & \textbf{86.96}
  & \textbf{11.99} & \textbf{84.22} & \textbf{13.33} & \textbf{85.34} 
  & \textbf{14.65} & \textbf{86.87} & \textbf{14.44} & \textbf{85.69} & \textbf{15.68} & \textbf{86.89} & \textbf{14.79} & \textbf{86.30} \\

 Dict+Title
  & 9.51 & 83.33 & 6.15 & 82.04 & 6.78 & 82.49 & 11.80 & 84.68
  & 7.96 & 83.44 & 5.55 & 82.27
  & 5.37 & 84.13 & 3.85 & 83.26 & 5.34 & 84.23 & 4.23 & 83.73 \\

 Statis+Title
  & 9.22 & 82.89 & 8.66 & 83.51 & 8.48 & 83.25 & 9.07 & 83.21
  & 8.77 & 83.61 & 8.64 & 83.43 
  & 9.84 & 84.43 & 8.86 & 83.91 & 10.46 & 84.35 & 9.51 & 84.16 \\

 Dict+Statis+Title
  & 9.16  & 82.84 & 10.19  & 84.33 & 9.50 & 83.94 & 10.79 & 84.12
  & 8.38  & 82.90  & 10.00 & 84.19 
  & 10.88  & 85.28  & 9.37 & 84.22 & 11.91 & 85.29 & 10.38 & 84.77 \\

 Dict+StatisT+Title
  & 8.18 & 82.47 & 8.87 & 83.23 & 8.95 & 83.04 & 10.79 & 84.06  
  & 6.72 & 81.08 & 8.92 & 83.17
  & 10.44 & 85.15  & 9.89 & 84.04 & 10.71 & 84.97 & 10.23 & 84.56 \\

\midrule
\textbf{internVL-2.5} \\

 Title
  & 13.80 & 84.33 & 13.55 & 85.02 & 12.59 & 84.58 & \textbf{15.74} & 85.59
  & 13.34 & 84.46 & 13.38 & 84.91 
  & 10.50 & 85.17 & 9.58 & 84.24 & 11.28 & 85.18 & 10.22 & 84.71 \\

 Dict+Title
  & \textbf{16.15} & 85.54 & \textbf{15.69} & \textbf{86.02} & \textbf{14.80} & \textbf{85.62} & 15.73 & \textbf{86.34}
  & \textbf{15.22} & \textbf{85.87} & 9.09 & 85.92 
  & 9.58 & \textbf{86.29} & 7.00 & 85.14 & 9.69 & \textbf{86.44} & 7.91 & 85.76 \\
  
 Statis+Title
  & 13.79 & 84.68 & 13.22 & 85.65 & 13.04 & 85.39 & 12.90 & 85.70
  & 12.80 & 85.44 & 13.17 & 85.57 
  & 13.05 & 85.82 & 11.65 & 84.98 & 13.14 & 85.75 & 12.38 & 85.39 \\

 Dict+Statis+Title
  & 13.86 & 85.06 & 14.17 & 85.95 & 13.67 & 85.58 & 14.32 & 86.31
  & 13.28 & 85.23 & 14.04 & \textbf{85.85} 
  & 13.43 & 86.20 & 11.96 & 85.18 & 14.04 & 86.26 & 12.85 & 85.71 \\

 Dict+StatisT+Title
  & 14.74 & \textbf{85.66} & 14.30 & 85.88 & 13.68 & 85.55 & 15.04 & 86.00  
  & 13.14 & 84.76 & \textbf{14.17} & 85.79
  & \textbf{13.96} & 86.23 & \textbf{12.49} & \textbf{85.19} & \textbf{15.10} & \textbf{86.44} & \textbf{13.52} & \textbf{85.78} \\

\midrule
\textbf{qwen2.5-VL-3B} \\

 Title
  & 14.91 & \textbf{85.86} & \textbf{16.22} & \textbf{86.66} & \textbf{14.74} & \textbf{85.91} & 18.38 & \textbf{87.56}
  & 14.88 & \textbf{86.12} & \textbf{15.88} & \textbf{86.49} 
  & \textbf{17.78} & \textbf{87.30} & \textbf{16.39} & \textbf{86.19} & \textbf{18.98} & \textbf{87.31} & \textbf{17.40} & \textbf{86.76} \\
  
 Dict+Title
  & \textbf{15.70} & 85.76 & 15.61 & 86.00 & 14.35 & 85.50 & \textbf{19.53} & 87.51
  & \textbf{15.22} & 85.36 & 8.09 & 85.91
  & 9.38 & 86.75 & 6.55 & 85.71 & 9.86 & 86.81 & 7.46 & 86.25 \\

 Statis+Title
  & 13.64 & 85.04 & 14.48 & 85.98 & 13.39 & 85.44 & 17.57 & 87.14 
  & 13.45 & 85.42 & 14.28 & 85.86 
  & 14.19 & 86.63 & 13.68 & 85.68 & 14.71 & 86.62 & 14.06 & 86.16 \\

 Dict+Statis+Title
  & 13.91 & 85.26 & 14.00 & 85.83 & 12.73 & 85.36 & 17.16 & 86.95 
  & 13.32 & 85.15 & 13.77 & 85.73 
  & 14.15 & 86.74 & 13.24 & 85.60 & 15.19 & 86.76 & 13.94 & 86.19 \\

 Dict+StatisT+Title
  & 13.25 & 85.28 & 14.47 & 85.98 & 13.03 & 85.47 & 18.07 & 87.02 
  & 13.06 & 85.00 & 14.19 & 85.86 
  & 14.78 & 86.79 & 13.36 & 85.43 & 15.96 & 86.87 & 14.36 & 86.15 \\
  
      \bottomrule
    \end{tabular}%
  }
  \caption{Ablation study results for different models regarding data used from Pew and VisText datasets evaluated on F1 scores of \textbf{R}OUGE-\textbf{L} and \textbf{B}ERT\textbf{S}core scores.}
  \label{tab:ablation_rouge_bert}
\end{table*}

\noindent\textbf{PoT Effectiveness Influenced by Input Textual Data.}
While the evaluation results remain influenced by the underlying VLM performance, we observed that in over half of the cases, the combination of the title and Python dictionaries outperformed using the title alone. We attribute this to the fact that directly extracted data, despite potential noise, can retain more valuable information than purely generated text, potentially steering the model toward more accurate outputs. However, this also highlights the power of using the PoT strategy, as it guides the model to emphasize more on the enhanced inaccuracies and noise with the poorly extracted data, while weakening the chart analysis, which negatively impacts the overall performance of the model pipeline.
In addition, we observed that the PoT strategy can consistently outperform in most cases with the InternVL model. This indicates the effectiveness of the PoT strategy with a pretrained VLM, which is better than directly using the title to enhance the overall pipeline performance in the chart summarization.

\section{Future Work}

From experimental results, it is observed that the summarization with the PoT strategy varied by different types of charts that the model was captioning. Most models performed well on relatively simpler bar and pie charts, while struggling with more complex charts, such as multiple line or scatter plots. This indicates that the generalizability requirement of the summarization task may involve some sort of normalization or some way to bridge the gap between the varying levels of complexity presented by the chart.
%
%
In future work, we would like to explore more sophisticated PoT approaches capable of generating longer and richer statistical information during the pipeline, thereby enhancing the quality of chart summaries. Since the PoT strategy in this work only extends outputs from short, answer-like responses to relatively concise statistical dictionaries. However, for the chart summarization task, we believe the PoT strategy contains untapped potential to capture factual numeric data by its statistical reasoning capability.
Moreover, given the significant influence of the PoT-generated information in model inference, we will also further investigate whether the PoT can contribute to mitigating hallucination errors in the chart summarization process, improving the overall factual accuracy of generated chart summaries.

\section{Conclusion}
In this work, we conducted a systematic evaluation of the Program-of-Thought (PoT) prompting strategy across currently used lightweight vision-language models under the zero-shot settings on the Pew and VisText benchmarks for the chart summarization task. Our experiments reveal that the efficacy of the PoT varies markedly with model architectures and sizes, pretrained data, corresponding to types of charts, including area, bar, line, pie, and scatter.
In conditions of VLMs and chart types, the PoT proved to be a competitive alternative to the Direct and MCoT prompting approaches with pretrained model, such as the InternVL.
Beyond prompting strategies, we introduced a novel chart-to-dictionary auxiliary task, demonstrating its promise for capturing robust and semantic nuances in chart understanding, which is also conveniently applicable with the PoT. As charts grow more complex along with the data they represent, there is a need to establish a data structure to evaluate chart-parsing outside the table due to data loss that occurs from the chart to the table.

\section*{Limitations}
The diverse performance of the PoT strategy across the evaluated models raises several important considerations. The model architecture and size likely play a significant role in determining the effectiveness of different prompting strategies. The models used in this paper were of lightweight VLMs. While effective in the presented lightweight models, the language decoder may have yielded too low conclusive powers on the efficacy of the PoT and CoT prompting methods relative to direct prompting. However, it is seen that the PoT strategy still can offer comparable results to the other prompting methodologies using lightweight VLMs in some cases or for some chart types, which indicates that on higher parameter models, it can be assumed that, in the worst case, these different prompting techniques may offer similar results.
The research design, comparing three zero-shot prompting methods across four distinct vision-language models and a set of tasks, provides a valuable initial exploration of the PoT's potential on chart summarization with VLMs. Further research can implement few-shot reasoning with examples that can hypothetically increase performance. Additionally, the study focused its experimentation on lightweight VLMs, which might have contributed to the poor results in text generation. Expanding the scope of the study to larger parameter models might lead to more conclusive results.


\section*{Ethics Statement}
To the best of the researchers' knowledge, all datasets used in this study were sourced from publicly available benchmarks. The authors of the benchmark dataset also have obtained the license to distribute the dataset for non-malicious purposes intent which this research has abided by.

\bibliography{custom}

\appendix

\section{Experiment Set-up}
\label{sec:apix_experiment}
The experiments are conducted with loaded pre-trained models from the vLLM API. As much as possible, the default parameters were used, unless suggested otherwise from official documentation. The temperature is set to 0.2, and the repetition penalty is set to 1.2 across all runs. All experiments are carried out on our machine (CPU: Intel(R) Core(TM) i9-9920X CPU @ 3.50GHz, GPU: 1 NVIDIA RTX3090). Python code generation for producing statistics by the Qwen2.5-Coder-14B-Instruct model is the most computationally costly task, which costs 10-12 hours on 1 GPU.

\section{Extended Evaluation Details}
\label{sec:evaluation_analysis}
\subsection{Dataset Analysis}
\label{sec:apix_dataset}

\begin{figure*}[t]
    \centering
    \includegraphics[width=0.95\linewidth]{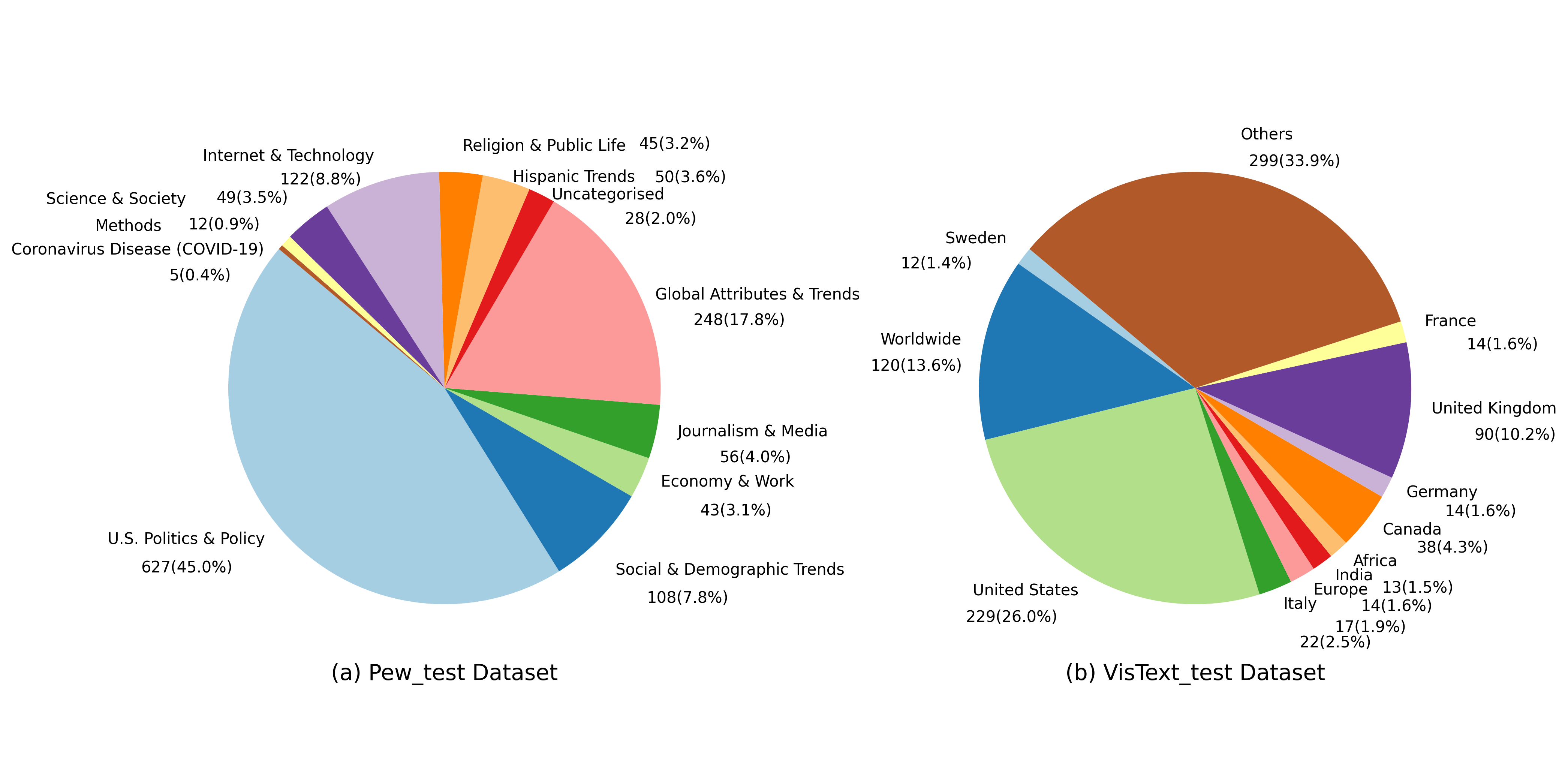}
    \caption{The distributions of topics of VisText and Pew test datasets.}
    \label{fig:dist_topics}
\end{figure*}

We chose the Pew \cite{kantharaj2022chart} (GPL-3.0 license) and VisText \cite{tang2023vistext} (GPL-3.0 license) large-domain English datasets to investigate and evaluate our PoT strategy for generating L2/L3 content in chart summarization, as they provide rich and suitable L2/L3 captions for this task. The VisText is built upon the Statista \cite{kantharaj2022chart} dataset, but with additionally detailed labelled L2/L3 captions.
Since the chart labelled in the VisText may have multiple L2/L3 captions, we automatically selected the longest L2/L3 captions in the test set of the VisText dataset as gold summaries paired to charts for the chart summarization task. The statistics of the Pew and VisText datasets used in this paper are presented in Table~\ref{tab:dataset_statis}.
%
%
In addition, the distribution of topics covered in the Pew and VisText datasets is illustrated in Figure~\ref{fig:dist_topics}.


\begin{table}[h]
\centering
\resizebox{\linewidth}{!}{%
\begin{tabular}{lcccccc}
\toprule
\multirow{2}{*}{\textbf{Statistic}}  & \multicolumn{3}{c}{\textbf{Pew}} & \multicolumn{3}{c}{\textbf{VisText}} \\
\cmidrule(lr){2-7}
      & Simp. & Comp. & All & Simp. & Comp. & All \\
\midrule
\#Vocab.        & 3,529   & 8,342  & 9,342  & 3,413  & 1,995  & 4,360   \\
Avg.Character   & 454     & 522    & 511    & 165    & 152    & 161     \\
Avg.Token       & 91      & 106    & 104    & 34     & 31     & 33      \\
Avg.Sentence    & 2.86    & 3.33   & 3.26   & 1.16   & 0.99   & 1.11    \\
\bottomrule
\end{tabular}
}
\caption{Statistics of datasets by \textbf{Simp}le and \textbf{Comp}lex complexities of the Pew and VisText test sets.}
\label{tab:dataset_statis}
\end{table}

\subsection{Experiment Implementations}
We mainly used DeepSeek-VL2 (deepseek-VL2-tiny) \cite{wu2024deepseekvl2tiny} for testing and our experiments. Additionally, we also tested the following models: InternVL (internVL-2.5-4B) \cite{chen2024internvl}, LLaVA-NeXT (llava-v1.6-mistral-7b-hf) \cite{liu2023llavav16mistral7bhf}, and Qwen-2.5 (qwen2.5-VL-3B-Instruct) \cite{qwen2025qwen25vl} for main and ablation experiments.
InternVL was reported to have one of our benchmark datasets, VisText, in its pretraining datasets \cite{chen2024internvl}. This likely contributed to its stronger performance, highlighting the potential benefits of pretraining. For the other models, there is no overlap with our benchmark datasets, nor evidence suggesting that the models were semantically aligned with the test distributions or were familiar with recurring chart patterns from their source papers.
All experiments were done in Python 3.12 using the vLLM \cite{kwon2023vllm} library, with the models being implemented at the zero-shot setting.
Similar to the previous work, the usage of ‘\#’ tokens in the generated tokens was restricted to avoid the pitfalls of only generating the reasoning chain as comments instead of executable code.


\subsection{Evaluation Metric Descriptions}
\label{sec:apix_evaluation}
To quantitatively measure the performance of our proposed method in chart summarization, we employ two popular automatic evaluation metrics in chart understanding: BLEU (Bilingual Evaluation Understudy) and CIDEr (Consensus-based Image Description Evaluation), in addition to two also well-known automatic evaluation metrics in text summarization: ROUGE \cite{lin2004rouge} and BERTScore \cite{zhang2020bertscore}. In order to evaluate factual correctness in chart summarization, we additionally adopt UniEval \cite{zhong2022unieval} and AlignScore \cite{zha2023alignscore}.

\noindent \textbf{BLEU} \cite{post2018callbleu} This score calculates the n-gram overlap between the ground-truth summary and the generated summary. It indicates lexical similarity between the generated and ground-truth text, assessing how closely the generated text replicates word sequences that occur in the reference.

\noindent \textbf{CIDEr} \cite{vedantam2015cider} This score measures the TFIDF weighted n-gram overlaps between reference and generated text. By weighting n-grams according to their value in a reference summary corpus, CIDEr seeks to more accurately capture the informativeness and relevance of generated descriptions, especially in image and chart captioning tasks.

BLEU and CIDEr are commonly used metrics throughout natural language generation, image captioning, and chart summarization. Together, they capture a more nuanced quantitative measure of model performance in terms of surface similarity and content alignment with reference summaries. While we note that reference-based measures like BLEU and CIDEr do have some limitations, since they can have loose correlation with human preference for aspects of semantic equivalence and factuality, their popularity and ability to provide an initial quantitative score make them effective measures in chart summarization model evaluation. As a result, we consider additional metrics for evaluating the chart summarization.


\noindent \textbf{ROUGE} \cite{lin2004rouge} This score is a prevailing metric in text summarization research based on semantic similarity.

\noindent \textbf{BERTScore} \cite{zhang2020bertscore} This score offers a complementary perspective by quantifying semantic similarity between system outputs and reference texts. 

\noindent \textbf{UniEval} \cite{zhong2022unieval} This score assesses factual correctness by framing the evaluation as a Boolean question-answering task.

\noindent \textbf{AlignScore} \cite{zha2023alignscore} This score evaluates factual correctness by quantifying the information alignment between two arbitrary text pieces.

\section{Prompts}
\label{sec:apix_prompts}

\subsection{LM Chat Templates}
\label{sec:apix_prompts_chat_template}

We show an exemplar from our chat templates for internVL here. The full chat templates can be found in our repository.

\begin{lstlisting}
{%- set ns = namespace(found_image=false) -%}
{{ bos_token }}
{%- for message in messages %}
    {%- if message['role'] == 'system' %}
        {{- '<|im_start|>system\n' + message['content'] + '<|im_end|>\n' -}}
    {%- elif message['role'] == 'user' %}
        {%- set content = message['content'] -%}
        {%- if '<image>' in content and not ns.found_image %}
            {%- set content = content | replace('<image>', '<image>\n', 1) -%}
            {%- set ns.found_image = true -%}
        {%- endif -%}
        {{- '<|im_start|>user\n' + content + '<|im_end|>\n' -}}
    {%- elif message['role'] == 'assistant' %}
        {{- '<|im_start|>assistant\n' + message['content'] + '<|im_end|>\n' -}}
    {%- endif %}
{%- endfor %}
{%- if add_generation_prompt %}
    {{- '<|im_start|>assistant\n' -}}
{%- endif %}
\end{lstlisting}

\subsection{Chart-to-Dictionary Extraction with Program of Thoughts}
\label{sec:apix_prompts_ctd}

Similar to the chart-to-table task, this is done in a zero-shot setting. We employ the core concept of PoT to guide the VLM in generating a valid and executable Python dictionary from the input chart.

\begin{lstlisting}
user_prompt = "<img_placeholder>\nConvert the chart into a python dictionary `chart_dict`. Only consider the chart's data when summarizing."
assistant_ = "```python\n chart_dict ="
\end{lstlisting}

We discover that with the request message of \textbf{"check errors"} within the prompt, the LM can implicitly check and correct both syntax errors in the output format and the facts in the data.

\begin{lstlisting}
user_prompt = "<img_placeholder>\nConvert the chart into a python dictionary `chart_dict`. Check json syntax errors. Only consider the chart's data when summarizing, no punctuations. Only return the valid version."
\end{lstlisting}

\subsection{Dictionary-to-Statistics with Program of Thoughts}
\label{sec:apix_prompts_pot}

The illustrated prompt content is the same used in VLMs tested in this work, but formatted specifically with each VLM's template.

\begin{lstlisting}
system_prompt = "You are a data analyst. You are given a dictionary that represents a chart called `chart_dict`. \
You need to implement the function `get_summary_statistics(chart_dict)` that takes the dictionary as input and returns a dictionary with the relevant statistics that can be used to summarize the chart. \
Avoid sorting dictionary objects directly and USE ONLY PYTHON BUILT-IN FUNCTIONS. Name the keys of the dictionary to elaborate how it is a descriptive statistic. When writing Python, follow the PEP style guide. \
Return ONLY the code of the function that will run without any errors and can work using `eval()`."

user = "Implement the function `get_summary_statistics` that takes a dictionary as input and returns a dictionary with the relevant statistics that can be used to summarize the chart using only built-in Python functions. Make sure to label the keys of the `summary_dict` to be descriptive The input dictionary is defined as {chart_dict}."

assistant_ = "```python\ndef get_summary_statistics(chart_dict):\n    # Define output dictionary `summary_dict` to store the summary statistics\n"
\end{lstlisting}

\subsection{Chart-to-Summary with Program of Thoughts}
\label{sec:apix_prompts_summ}

\begin{lstlisting}
user = "Summarize the insights of the chart with title: '{title}'. The summary use language similar to the chart. Don't explicitly describe chart elements such as chart type. NEVER START A SENTENCE WITH A NUMBER. The chart has the dictionary: {dictionary_str} and the summary_statistics: {summary_dict}."

assistant_ = "Let's think step by step to with as few steps as possible to summarize the chart: "
\end{lstlisting}

\section{Extended Results}

\subsection{Ablation Studies}
\label{sec:ablation_study_tests}
Table~\ref{tab:apix_text_ablation} and Table~\ref{tab:apix_ablation_rouge} present BLEU and CIDEr evaluation results, and ROUGE-1 and ROUGE-L evaluation results, respectively, for various VLMs, tested with different prompting strategies. Table~\ref{tab:apix_text_ablation_unieval_cohcons} and Table~\ref{tab:apix_text_ablation_unieval_flurel} present coherence, consistency, fluency, and relevance from UniEval evaluation results.
Table~\ref{tab:apix_ablation_textpye_aliuni}, Table~\ref{tab:apix_ablation_textpye_aliunicohcon}, and Table~\ref{tab:apix_ablation_textpye_aliuniflurel} present AlignScore-large score and UniEval score results for various VLMs regarding textual data types with our PoT chart summarization pipeline. We discover that even the evaluation scores of template-based statistics generation mostly outperform other PoT-based statistics generation, but most summaries generated by template-based statistics are low-quality. The factual evaluation results can also indicate that the simple rule-based transformation from a Python dictionary to a structured key-value statistical analysis is rigid and incurs higher computational and development costs.


With empirical results, how a candidate task to represent charts structurally can be an effective auxiliary to the existing chart-to-table task can be sort of answered. While the evaluation of chart-to-table might be more objective in its evaluation, there might be merit to exploring the chart-to-dictionary task for chart understanding. Not only this enable the integration of the chart in a PoT context, but it also facilitates a more robust representation of the chart, given the increasing complexities of charts in the wild. This work acknowledges that there is an overlap between chart redrawing and this task, but the chart redrawing tends to focus more on the reconstruction of the chart with executable matplotlib code rather than capturing the semantic nuances of the chart elements explored in this work.

\subsection{Comparison between Manual Template-based and PoT-based Statistics Generation}
\label{sec:apix_mannual_comp}

We show an exemplar of the predefined extracting data rules in our manual template method. The full rules can be found in our repository.

\begin{lstlisting}
if isinstance(values, list) and values and all(isinstance(x, (int, float)) for x in values):
    return [{
        "Category": prefix,
        "Total": len(values),
        "Sum": sum(values),
        "Average": statistics.mean(values),
        "Minimum": min(values),
        "Maximum": max(values),
        "Range": max(values) - min(values)
    }]
\end{lstlisting}

\begin{figure}[h]
\centering
  \includegraphics[width=0.9\linewidth]{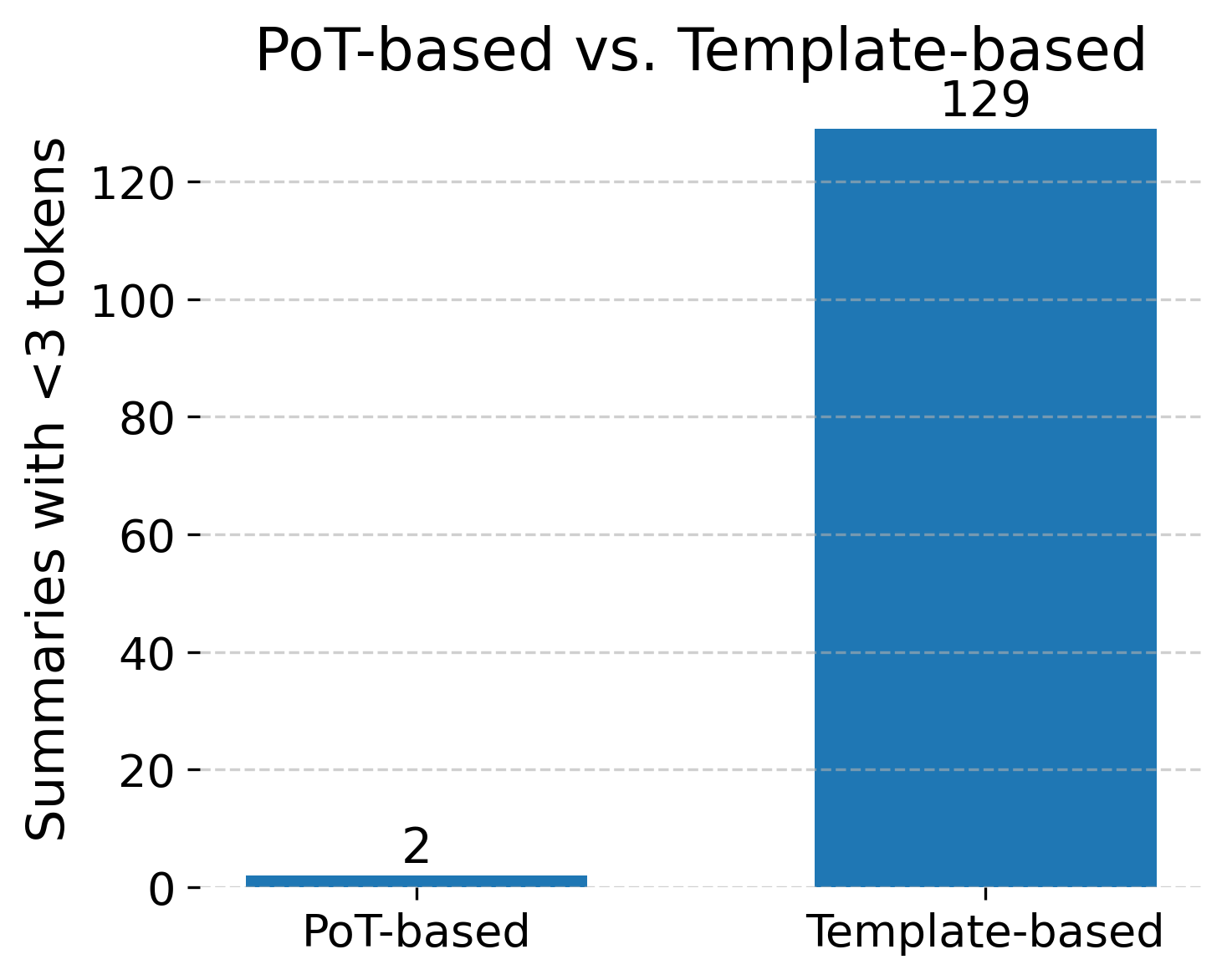}
  \caption{Histogram comparing the numbers of failure cases (output summaries $<$3 tokens in length) in the generated summaries from PoT-based and Template-based DeepSeek on the Pew dataset.}
  \label{fig:count_short_lines}
\end{figure}

Figure~\ref{fig:count_short_lines} shows a comparison of the numbers of short failed summaries generated by using the PoT-based and Template-based Deepseek model on the Pew dataset, indicating the effectiveness of using the PoT instead of a simple rule-based template with the VLM in chart summarization.

\begin{figure}[h]
  \includegraphics[width=\linewidth]{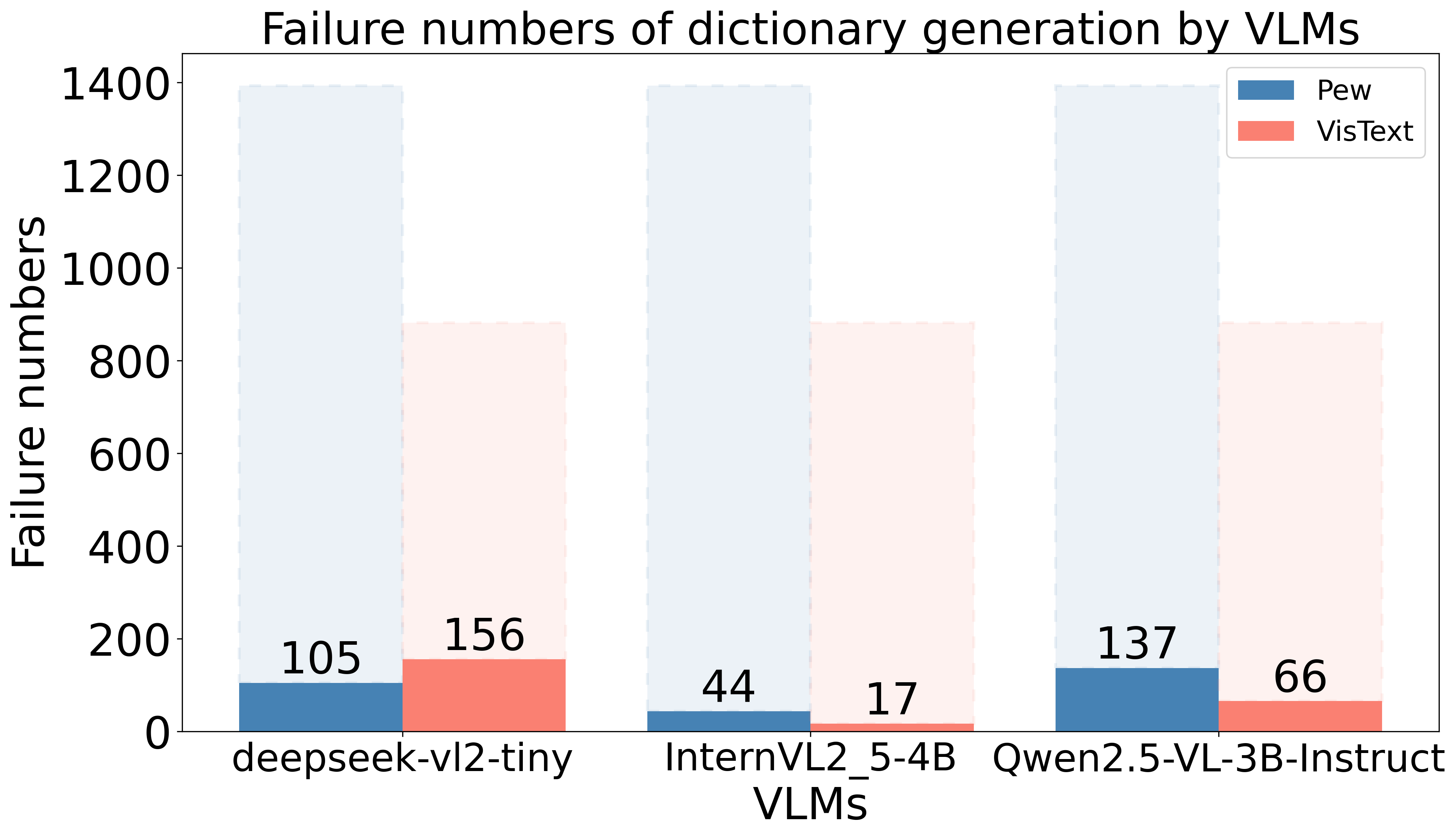}
  \caption{Histogram comparing the numbers of failure cases in the chart data dictionary generation by each VLM on each dataset.}
  \label{fig:dict_failurenum}
\end{figure}

\begin{table*}[t]
  \centering
  \renewcommand{\arraystretch}{1.2}
  \resizebox{\textwidth}{!}{%
    \begin{tabular}{ll*{20}{r}}
      \toprule
      \multirow{3}{*}{\shortstack{\textbf{VLM} \\  \\ \textbf{+Textual Data}}}
        & \multicolumn{12}{c}{\textbf{Pew}}
        & \multicolumn{8}{c}{\textbf{VisText}} \\
      \cmidrule(lr){2-13}\cmidrule(lr){14-21}
      &
         \multicolumn{2}{c}{Area}
        & \multicolumn{2}{c}{Bar}
        & \multicolumn{2}{c}{Line}
        & \multicolumn{2}{c}{Pie}
        & \multicolumn{2}{c}{Scatter}
        & \multicolumn{2}{c}{All}
        & \multicolumn{2}{c}{Area}
        & \multicolumn{2}{c}{Bar}
        & \multicolumn{2}{c}{Line}
        & \multicolumn{2}{c}{All} \\
      &
         BLEU & CIDEr
        & BLEU & CIDEr
        & BLEU & CIDEr
        & BLEU & CIDEr
        & BLEU & CIDEr
        & BLEU & CIDEr
        & BLEU & CIDEr
        & BLEU & CIDEr
        & BLEU & CIDEr
        & BLEU & CIDEr \\
      \midrule
      \textbf{deepseek-vl2-tiny} \\

 Title
  & \textbf{1.9682} & \textbf{0.0427} & \textbf{2.6653} & \textbf{0.0608} & \textbf{1.7169} & \textbf{0.0471} & \textbf{4.5805} & \textbf{0.1391} 
  & \textbf{0.7646} & 0.0412 & \textbf{2.4676} & \textbf{0.0591} 
  & \textbf{1.8347} & \textbf{0.0920} & \textbf{1.5262} & 0.0731 & \textbf{2.0429} & \textbf{0.0851} & \textbf{1.7346} & \textbf{0.0824} \\

 Dict+Title
  & 0.3425 & 0.0000 & 0.2343 & 0.0040 & 0.1940 & 0.0055 & 0.7802 & 0.0095 
  & 0.3621 & 0.0002 & 0.1707 & 0.0025 
  & 0.2472 & 0.0115 & 0.0853 & 0.0077 & 0.2067 & 0.0124 & 0.0855 & 0.0081 \\

 Statis+Title  
  & 0.3627 & 0.0182 & 0.3498 & 0.0214 & 0.2932 & 0.0118 & 0.2772 & 0.0133 
  & 0.5526 & \textbf{1.2809} & 0.3407 & 0.0183 
  & 0.4872 & 0.0688 & 0.5449 & 0.0582 & 0.4168 & 0.0654 & 0.5153 & 0.0636 \\
  
 Dict+Statis+Title
  & 0.6960 & 0.0135 & 0.6807 & 0.0236 & 0.6517 & 0.0251 & 0.6614 & 0.0341 
  & 0.3309 & 0.0011 & 0.6875 & 0.0235 
  & 0.5583 & 0.0713 & 0.4584 & \textbf{0.0737} & 1.1808 & 0.0796 & 0.6914 & 0.0754 \\

 Dict+StatisT+Title
  & 0.7589 & 0.0023 & 0.4311 & 0.0170 & 0.5564 & 0.0181 & 0.3408 & 0.0320 
  & 0.3350 & 0.0279 & 0.4914 & 0.0173 
  & 0.4408 & 0.0676 & 0.7812 & 0.0568 & 0.8565 & 0.0538 & 0.7502 & 0.0589 \\
  
\midrule
\textbf{internVL-2.5} \\

 Title
  & 3.6507 & 0.0426 & \textbf{3.5832} & 0.0318 & \textbf{2.7521} & 0.0296 & \textbf{4.6431} & 0.1025 
  & 2.6224 & 0.0001 & \textbf{3.4041} & 0.0328 
  & 1.1306 & 0.0125 & 0.9387 & 0.0088 & 1.3401 & 0.0212 & 1.0808 & 0.0130 \\
  
 Dict+Title  
  & 3.7973 & 0.1391 & 3.1843 & 0.0650 & 2.2829 & 0.0612 & 2.7083 & 0.1088
  & 1.6723 & 0.0569 & 0.7148 & 0.0052 
  & 0.2476 & 0.0057 & 0.0790 & 0.0023 & 0.4311 & 0.0110 & 0.2141 & 0.0047 \\

 Statis+Title  
  & 3.2816 & 0.0253 & 2.0090 & 0.0569 & 1.9310 & 0.0478 & 1.6361 & 0.0591 
  & 1.0602 & 0.0443 & 1.9939 & 0.0540 
  & 1.2121 & 0.0100 & 1.0379 & 0.0169 & 1.5626 & 0.0210 & 1.2156 & 0.0157 \\
  
 Dict+Statis+Title 
  & 3.6093 & 0.1211 & 3.1860 & \textbf{0.0697} & 2.5661 & 0.0615 & 2.9342 & 0.1188 
  & 1.8525 & \textbf{0.0960} & 3.0319 & \textbf{0.0695} 
  & 1.4938 & 0.0326 & 1.0729 & 0.0102 & 1.9497 & 0.0237 & 1.3735 & 0.0192 \\

 Dict+StatisT+Title 
  & \textbf{4.1720} & \textbf{0.1772} & 3.1456 & 0.0633 & 2.4598 & \textbf{0.0770} & 2.7016 & \textbf{0.1286} 
  & \textbf{3.2064} & 0.0431 & 3.0008 & 0.0689
  & \textbf{1.7194} & \textbf{0.0555} & \textbf{1.2597} & \textbf{0.0205} & \textbf{2.3460} & \textbf{0.0506} & \textbf{1.5926} & \textbf{0.0371} \\
  
\midrule
\textbf{qwen2.5-VL-3B} \\

 Title
  & 1.9350 & 0.0523 & \textbf{3.6251} & \textbf{0.1002} & \textbf{2.5562} & 0.0643 & \textbf{5.9420} & 0.1384 
  & 2.0714 & 0.0272 & \textbf{3.3929} & \textbf{0.0905} 
  & \textbf{2.6399} & \textbf{0.1481} & \textbf{2.1772} & \textbf{0.0979} & \textbf{3.1147} & \textbf{0.1519} & \textbf{2.4984} & \textbf{0.1254} \\
  
 Dict+Title
  & 2.6846 & \textbf{0.0953} & 3.1135 & 0.0693 & 2.2941 & 0.0652 & 3.6053 & \textbf{0.1937} 
  & 1.5115 & 0.0629 & 0.6707 & 0.0060 
  & 0.3687 & 0.0168 & 0.0869 & 0.0090 & 0.4078 & 0.0136 & 0.1515 & 0.0097 \\

 Statis+Title
  & \textbf{3.3383} & 0.0409 & 3.3091 & 0.0734 & 2.3678 & 0.0597 & 3.8250 & 0.1662 
  & 1.0761 & 0.0203 & 3.0906 & 0.0712 
  & 1.6593 & 0.0780 & 1.4806 & 0.0801 & 2.0928 & 0.0890 & 1.6639 & 0.0826 \\
  
 Dict+Statis+Title
  & 3.0823 & 0.0830 & 3.0102 & 0.0727 & 2.1315 & 0.0616 & 3.3978 & 0.1346 
  & 2.0385 & 0.0294 & 2.8237 & 0.0711 
  & 1.6373 & 0.0781 & 1.2874 & 0.0596 & 2.1118 & 0.0714 & 1.5484 & 0.0678 \\

 Dict+StatisT+Title
  & 2.4238 & 0.0222 & 3.2131 & 0.0640 & 2.2744 & \textbf{0.0693} & 3.4002 & 0.1018 
  & 2.6648 & \textbf{0.0662} & 2.9969 & 0.0652 
  & 1.7080 & 0.1080 & 1.4815 & 0.0688 & 2.3149 & 0.1042 & 1.6950 & 0.0883 \\
      \bottomrule
    \end{tabular}%
  }
  \caption{Ablation study results (BLEU / CIDEr) for different models regarding data used from Pew and VisText datasets.}
  \label{tab:apix_text_ablation}
\end{table*}

\begin{table*}[t]
  \centering
  \renewcommand{\arraystretch}{1.2}
  \resizebox{\textwidth}{!}{%
    \begin{tabular}{ll*{20}{r}}
      \toprule
      \multirow{3}{*}{\shortstack{\textbf{VLM} \\  \\ \textbf{+Textual Data}}}
        & \multicolumn{12}{c}{\textbf{Pew}}
        & \multicolumn{8}{c}{\textbf{VisText}} \\
      \cmidrule(lr){2-13}\cmidrule(lr){14-21}
      &
        \multicolumn{2}{c}{Area}
        & \multicolumn{2}{c}{Bar}
        & \multicolumn{2}{c}{Line}
        & \multicolumn{2}{c}{Pie}
        & \multicolumn{2}{c}{Scatter}
        & \multicolumn{2}{c}{All}
        & \multicolumn{2}{c}{Area}
        & \multicolumn{2}{c}{Bar}
        & \multicolumn{2}{c}{Line}
        & \multicolumn{2}{c}{All} \\
      &
        R-1 & R-L
        & R-1 & R-L
        & R-1 & R-L
        & R-1 & R-L
        & R-1 & R-L
        & R-1 & R-L
        & R-1 & R-L
        & R-1 & R-L
        & R-1 & R-L
        & R-1 & R-L \\
      \midrule
      \textbf{deepseek-vl2-tiny} \\

 Title
  & \textbf{24.62} & \textbf{13.57} & \textbf{25.88} & \textbf{13.57} & \textbf{23.66} & \textbf{12.26} & \textbf{29.17} & \textbf{16.98}
  & \textbf{24.03} & \textbf{11.99} & \textbf{25.40} & \textbf{13.33} 
  & \textbf{22.37} & \textbf{14.65} & \textbf{21.72} & \textbf{14.44} & \textbf{23.56} & \textbf{15.68} & \textbf{22.33} & \textbf{14.79} \\

 Dict+Title
  & 14.82 & 9.51 & 8.94 & 6.15 & 9.97 & 6.78 & 16.71 & 11.80
  & 13.05 & 7.96 & 8.05 & 5.55 
  & 7.35 & 5.37 & 4.98 & 3.85 & 7.31 & 5.34 & 5.56 & 4.23 \\

 Statis+Title
  & 15.23 & 9.22 & 13.65 & 8.66 & 13.12 & 8.48 & 12.66 & 9.07
  & 14.72 & 8.77 & 13.53 & 8.64 
  & 14.26 & 9.84 & 12.21 & 8.86 & 14.70 & 10.46 & 13.38 & 9.51 \\

 Dict+Statis+Title
  & 15.86  & 9.16   & 16.70  & 10.19 & 16.32 & 9.50 & 16.81 & 10.79
  & 13.98  & 8.38   & 16.57 & 10.00 
  & 16.48  & 10.88  & 13.75 & 9.37 & 17.14 & 11.91 & 15.29 & 10.38 \\

 Dict+StatisT+Title
  & 15.24 & 8.18 & 14.19 & 8.87 & 15.34 & 8.95 & 16.48  & 10.79  & 10.64  & 6.72 & 14.53  & 8.92
  & 15.46  & 10.44  & 14.44 & 9.89 & 16.12 & 10.71 & 15.11 & 10.23 \\

\midrule
\textbf{internVL-2.5} \\

 Title
  & 27.44 & 13.80 & \textbf{28.86} & 13.55 & 26.81 & 12.59 & \textbf{30.08} & \textbf{15.74}
  & 27.82 & 13.34 & \textbf{28.37} & 13.38 
  & 17.17 & 10.50 & 16.19 & 9.58 & 18.21 & 11.28 & 16.92 & 10.22 \\
  
 Dict+Title
  & \textbf{28.78} & \textbf{16.15} & 28.52 & \textbf{15.69} & 25.93 & \textbf{14.80} & 27.67 & 15.73
  & 27.57 & \textbf{15.22} & 15.53 & 9.09 
  & 15.87 & 9.58 & 10.69 & 7.00 & 15.53 & 9.69 & 12.43 & 7.91 \\

 Statis+Title
  & 25.52 & 13.79 & 24.53 & 13.22 & 24.03 & 13.04 & 21.58 & 12.90
  & 24.05 & 12.80 & 24.33 & 13.17 
  & 20.81 & 13.05 & 18.58 & 11.65 & 20.81 & 13.14 & 19.72 & 12.38 \\

 Dict+Statis+Title
  & 26.64 & 13.86 & 28.52 & 14.17 & \textbf{27.27} & 13.67 & 26.11 & 14.32
  & 27.55 & 13.28 & 28.11 & 14.04 
  & 22.24 & 13.43 & 20.30 & 11.96 & 22.66 & 14.04 & 21.38 & 12.85 \\

 Dict+StatisT+Title
  & 26.86 & 14.74 & 28.18 & 14.30 & 26.66 & 13.68 & 26.23 & 15.04  & \textbf{28.19} & 13.14 & 27.73 & \textbf{14.17}
  & \textbf{22.52} & \textbf{13.96} & \textbf{20.53} & \textbf{12.49} & \textbf{23.43} & \textbf{15.10} & \textbf{21.76} & \textbf{13.52} \\

\midrule
\textbf{qwen2.5-VL-3B} \\
 Title
  & 24.83 & 14.91 & \textbf{30.29} & \textbf{16.22} & \textbf{27.70} & \textbf{14.74} & \textbf{32.16} & 18.38
  & \textbf{29.51} & 14.88 & \textbf{29.62} & \textbf{15.88} 
  & \textbf{26.14} & \textbf{17.78} & \textbf{24.85} & \textbf{16.39} & \textbf{27.12} & \textbf{18.98} & \textbf{25.74} & \textbf{17.40} \\
  
 Dict+Title
  & \textbf{26.12} & \textbf{15.70} & 27.49 & 15.61 & 25.49 & 14.35 & 30.70 & \textbf{19.53}
  & 29.14 & \textbf{15.22} & 13.50 & 8.09
  & 14.10 & 9.38 & 9.29 & 6.55 & 15.00 & 9.86 & 10.91 & 7.46 \\

 Statis+Title
  & 24.75 & 13.64 & 27.53 & 14.48 & 25.53 & 13.39 & 29.58 & 17.57 
  & 27.06 & 13.45 & 27.06 & 14.28 
  & 22.11 & 14.19 & 21.27 & 13.68 & 22.70 & 14.71 & 21.83 & 14.06 \\

 Dict+Statis+Title
  & 25.56 & 13.91 & 26.58 & 14.00 & 24.62 & 12.73 & 28.36 & 17.16 
  & 25.85 & 13.32 & 26.13 & 13.77 
  & 22.23 & 14.15 & 20.46 & 13.24 & 23.20 & 15.19 & 21.59 & 13.94 \\

 Dict+StatisT+Title
  & 23.72 & 13.25 & 27.44 & 14.47 & 25.24 & 13.03 & 29.28 & 18.07 
  & 25.69 & 13.06 & 26.89 & 14.19 
  & 22.10 & 14.78 & 20.74 & 13.36 & 23.70 & 15.96 & 21.82 & 14.36 \\
  
      \bottomrule
    \end{tabular}%
  }
  \caption{Ablation study results (ROUGE-1 / ROUGE-L) for different models regarding data used on Pew and VisText datasets.}
  \label{tab:apix_ablation_rouge}
\end{table*}

\begin{table*}[t]
  \centering
  \renewcommand{\arraystretch}{1.2}
  \resizebox{\textwidth}{!}{%
    \begin{tabular}{ll*{19}{r}}
      \toprule
         \multirow{3}{*}{\shortstack{\textbf{VLM} \\  \\ \textbf{-Prompting}}}
        & \multicolumn{12}{c}{\textbf{Pew}}
        & \multicolumn{8}{c}{\textbf{VisText}} \\
      \cmidrule(lr){2-13}\cmidrule(lr){14-21}
      &
          \multicolumn{2}{c}{Area}
        & \multicolumn{2}{c}{Bar}
        & \multicolumn{2}{c}{Line}
        & \multicolumn{2}{c}{Pie}
        & \multicolumn{2}{c}{Scatter}
        & \multicolumn{2}{c}{All}
        & \multicolumn{2}{c}{Area}
        & \multicolumn{2}{c}{Bar}
        & \multicolumn{2}{c}{Line}
        & \multicolumn{2}{c}{All} \\
      &
          coh & cons
        & coh & cons
        & coh & cons
        & coh & cons
        & coh & cons
        & coh & cons
        & coh & cons
        & coh & cons
        & coh & cons
        & coh & cons \\
      \midrule
      \textbf{deepseek-vl2-tiny} \\

   ZeroShot-Direct 
  & \textbf{86.91} & 48.42 & \textbf{80.57} & 55.94 & \textbf{85.51} & 52.75 & \textbf{85.14} & \textbf{65.36}
  & \textbf{84.82} & 71.25 & \textbf{82.08} & 55.48 
  & \textbf{87.91} & 69.66 & 82.15 & 60.05 & \textbf{87.84} & \textbf{69.94} & \textbf{85.08} & 65.03 \\

   ZeroShot-MCoT  
  & 78.14 & \textbf{52.97} & 77.88 & \textbf{59.61} & 76.82 & \textbf{58.31} & 82.94 & 62.99 
  & 80.91 & \textbf{72.62} & 77.80 & \textbf{59.42} 
  & 83.20 & \textbf{71.06} & \textbf{83.01} & \textbf{66.03} & 83.20 & 69.34 & 83.11 & \textbf{68.19} \\


   ZeroShot-PoT  
  & 44.51 & 32.56 & 47.73 & 43.95 & 44.33 & 40.39 & 48.31 & 45.93 
  & 43.74 & 41.96 & 46.81 & 42.93
  & 52.60 & 48.64 & 51.62 & 47.32 & 49.61 & 46.52 & 51.40 & 47.48 \\
\midrule
\textbf{internVL-2.5} \\

         ZeroShot-Direct 
  & 81.15 & \textbf{55.46} & 81.70 & 60.16 & 84.77 & 55.98 & \textbf{88.64} & \textbf{67.23}
  & 89.11 & 59.54 & 82.74 & 59.24 
  & 90.54 & 62.16 & 88.90 & 60.34 & 89.56 & 62.71 & 89.50 & 61.41 \\

        ZeroShot-MCoT  
  & \textbf{86.89} & 53.07 & 81.20 & 54.44 & 81.89 & 51.98 & 81.16 & 57.06 
  & 79.06 & 55.26 & 81.43 & 53.89 
  & 88.26 & 58.34 & 88.06 & 57.88 & 87.71 & 57.94 & 88.03 & 58.02 \\
  
         ZeroShot-PoT  
  & 86.79 & 47.76 & \textbf{90.84} & \textbf{61.34} & \textbf{90.55} & \textbf{59.63} & 87.45 & 58.23 
  & \textbf{94.53} & \textbf{67.90} & \textbf{90.65} & \textbf{60.70}
  & \textbf{93.17} & \textbf{65.92} & \textbf{94.04} & \textbf{64.05} & \textbf{93.18} & \textbf{63.67} & \textbf{93.60} & \textbf{64.46} \\
\midrule
\textbf{llava-NeXT}  \\

         ZeroShot-Direct 
  & \textbf{91.08} & \textbf{60.18} & 89.18 & \textbf{62.70} & 92.15 & \textbf{60.67} & \textbf{86.85} & 66.79 
  & 81.96 & 67.66 & 89.80 & \textbf{62.33} 
  & \textbf{94.43} & \textbf{67.26} & \textbf{93.07} & \textbf{67.23} & \textbf{93.35} & \textbf{68.33} & \textbf{93.50} & \textbf{67.50} \\

          ZeroShot-MCoT  
  & 92.90 & 55.01 & $/$ & $/$ & 92.00 & 59.19 & 91.26 & \textbf{66.69} 
  & 93.73 & \textbf{66.05} & $/$ & $/$ 
  & $/$ & $/$ & 93.26 & 65.01 & $/$ & $/$ & $/$ & $/$ \\

          ZeroShot-PoT  
  & 74.14 & 44.92 & 88.52 & 62.23 & \textbf{87.72} & 59.42 & 87.54 & 62.48 
  & \textbf{90.63} & 62.64 & 88.10 & 61.29 
  & 92.91 & 64.74 & 91.55 & 65.75 & 92.83 & 64.20 & 92.23 & 65.10 \\
\midrule
\textbf{qwen2.5-VL-3B}  \\

       ZeroShot-Direct 
  & \textbf{87.39} & 53.39 & \textbf{88.87} & 66.98 & \textbf{87.14} & 58.37 & 91.00 & 72.48 
  & \textbf{94.42} & 78.91 & \textbf{88.54} & 64.92 
  & 87.62 & 72.22 & \textbf{86.76} & 64.40 & 87.17 & 71.31 & 87.09 & 68.18 \\

       ZeroShot-MCoT  
  & 80.06 & \textbf{61.37} & 86.08 & \textbf{67.24} & 85.49 & \textbf{60.40} & \textbf{91.11} & \textbf{74.43} 
  & 93.24 & \textbf{79.90} & 86.07 & \textbf{65.79} 
  & \textbf{88.42} & \textbf{74.59} & 86.04 & \textbf{67.87} & \textbf{89.18} & \textbf{73.11} & \textbf{87.44} & \textbf{70.95} \\
  
       ZeroShot-PoT  
  & 87.38 & 57.06 & 82.68 & 61.71 & 83.35 & 56.77 & 87.48 & 73.66 
  & 93.30 & 65.71 & 83.17 & 60.80 
  & 85.33 & 67.24 & 84.06 & 63.47 & 85.32 & 68.70 & 84.71 & 65.75 \\

      \bottomrule
    \end{tabular}%
  }
  \caption{Evaluation results of VLMs on different prompting methods on Pew and VisText datasets evaluated on UniEval-\textbf{coh}erence and UniEval-\textbf{cons}istency.}
  \label{tab:apix_text_ablation_unieval_cohcons}
\end{table*}

\begin{table*}[t]
  \centering
  \renewcommand{\arraystretch}{1.2}
  \resizebox{\textwidth}{!}{%
    \begin{tabular}{ll*{19}{r}}
      \toprule
         \multirow{3}{*}{\shortstack{\textbf{VLM} \\  \\ \textbf{-Prompting}}}
        & \multicolumn{12}{c}{\textbf{Pew}}
        & \multicolumn{8}{c}{\textbf{VisText}} \\
      \cmidrule(lr){2-13}\cmidrule(lr){14-21}
      &
          \multicolumn{2}{c}{Area}
        & \multicolumn{2}{c}{Bar}
        & \multicolumn{2}{c}{Line}
        & \multicolumn{2}{c}{Pie}
        & \multicolumn{2}{c}{Scatter}
        & \multicolumn{2}{c}{All}
        & \multicolumn{2}{c}{Area}
        & \multicolumn{2}{c}{Bar}
        & \multicolumn{2}{c}{Line}
        & \multicolumn{2}{c}{All} \\
      &
          flu & rel
        & flu & rel
        & flu & rel
        & flu & rel
        & flu & rel
        & flu & rel
        & flu & rel
        & flu & rel
        & flu & rel
        & flu & rel \\
      \midrule
      \textbf{deepseek-vl2-tiny} \\

   ZeroShot-Direct 
  & \textbf{95.08} & \textbf{85.59} & \textbf{90.91} & \textbf{78.78} & \textbf{92.45} & \textbf{83.72} & 91.45 & \textbf{84.22} 
  & \textbf{92.99} & \textbf{83.48} & \textbf{91.40} & \textbf{80.33}
  & \textbf{94.63} & \textbf{85.16} & \textbf{92.39} & 79.22 & \textbf{94.33} & \textbf{85.33} & \textbf{93.46} & \textbf{82.30} \\

   ZeroShot-MCoT  
  & 90.53 & 76.52 & 89.34 & 75.65 & 88.58 & 74.09 & \textbf{91.57} & 80.50 
  & 88.29 & 80.02 & 89.22 & 75.47 
  & 89.85 & 79.54 & 90.82 & \textbf{79.37} & 91.27 & 79.81 & 90.67 & 79.52 \\


   ZeroShot-PoT  
  & 86.10 & 41.17 & 85.39 & 45.03 & 83.07 & 41.92 & 87.04 & 45.75 
  & 87.47 & 40.56 & 84.89 & 44.17
  & 84.14 & 48.39 & 83.02 & 46.21 & 83.08 & 44.87 & 83.34 & 46.47 \\
\midrule
\textbf{internVL-2.5} \\

         ZeroShot-Direct 
  & 83.32 & 81.21 & 86.38 & 81.52 & 87.87 & 84.65 & 84.43 & \textbf{87.57} 
  & 91.29 & 89.71 & 86.71 & 82.57 
  & 93.57 & 89.73 & 92.17 & 88.12 & 92.86 & 89.33 & 92.72 & 88.85 \\

        ZeroShot-MCoT  
  & 91.80 & \textbf{86.53} & 89.28 & 80.57 & 89.57 & 80.97 & 89.49 & 80.47 
  & 87.92 & 77.49 & 89.38 & 80.72 
  & 92.84 & 87.23 & 92.64 & 87.07 & 92.75 & 87.10 & 92.72 & 87.12 \\
  
         ZeroShot-PoT  
  & \textbf{93.40} & 85.96 & \textbf{94.70} & \textbf{90.18} & \textbf{94.92} & \textbf{89.82} & \textbf{95.05} & 85.81
  & \textbf{95.81} & \textbf{93.59} & \textbf{94.76} & \textbf{89.93} 
  & \textbf{95.23} & \textbf{92.07} & \textbf{95.23} & \textbf{92.87} & \textbf{95.45} & \textbf{91.96} & \textbf{95.28} & \textbf{92.43} \\
\midrule
\textbf{llava-NeXT}  \\

         ZeroShot-Direct 
  & 95.02 & 90.77 & 93.04 & \textbf{88.81} & \textbf{94.73} & \textbf{91.48} & 90.92 & 86.19
  & 94.10 & 81.14 & \textbf{93.44} & \textbf{89.34} 
  & \textbf{95.55} & 87.73 & \textbf{94.94} & 92.11 & \textbf{95.73} & \textbf{92.71} & \textbf{95.29} & \textbf{92.68} \\

          ZeroShot-MCoT  
  & \textbf{95.30} & \textbf{92.53} & $/$ & $/$ & 94.60 & 91.46 & \textbf{94.33} & \textbf{90.54} 
  & \textbf{95.10} & \textbf{93.28} & $/$ & $/$ 
  & $/$ & $/$ & 94.55 & \textbf{92.34} & $/$ & $/$ & $/$ & $/$ \\

          ZeroShot-PoT  
  & 92.13 & 73.79 & \textbf{93.48} & 87.95 & 93.06 & 86.86 & 92.64 & 85.74 
  & 93.16 & 89.53 & 93.33 & 87.43 
  & 95.16 & \textbf{92.02} & 94.62 & 90.19 & 94.92 & 91.73 & 94.84 & 91.06 \\
\midrule
\textbf{qwen2.5-VL-3B}  \\

       ZeroShot-Direct 
  & \textbf{95.15} & 86.34 & \textbf{93.70} & \textbf{87.65} & 93.40 & \textbf{86.26} & \textbf{95.07} & \textbf{90.19} 
  & \textbf{95.46} & 92.74 & \textbf{93.71} & \textbf{87.41}
  & \textbf{95.06} & 84.77 & \textbf{93.11} & \textbf{84.42} & \textbf{94.73} & 84.32 & \textbf{94.03} & 84.49 \\

       ZeroShot-MCoT  
  & 92.87 & 79.32 & 93.22 & 84.46 & \textbf{93.71} & 84.37 & 94.88 & 89.78 
  & 95.14 & 92.21 & 93.41 & 84.60 
  & 93.79 & \textbf{85.66} & 92.72 & 83.82 & 94.61 & \textbf{86.69} & 93.47 & \textbf{85.01} \\
  
       ZeroShot-PoT  
  & 93.91 & \textbf{87.23} & 92.16 & 82.01 & 92.50 & 82.87 & 93.78 & 87.31 
  & 94.45 & \textbf{92.99} & 92.34 & 82.58
  & 93.69 & 83.03 & 93.08 & 81.69 & 93.85 & 82.52 & 93.43 & 82.25 \\

      \bottomrule
    \end{tabular}%
  }
  \caption{Evaluation results of VLMs on different prompting methods on Pew and VisText datasets evaluated on UniEval-\textbf{flu}ency and UniEval-\textbf{rel}evance.}
  \label{tab:apix_text_ablation_unieval_flurel}
\end{table*}

\begin{table*}[t]
  \centering
  \renewcommand{\arraystretch}{1.2}
  \resizebox{\textwidth}{!}{%
    \begin{tabular}{ll*{20}{r}}
      \toprule
      \multirow{3}{*}{\shortstack{\textbf{VLM} \\  \\ \textbf{+Textual Data}}}
        & \multicolumn{12}{c}{\textbf{Pew}}
        & \multicolumn{8}{c}{\textbf{VisText}} \\
      \cmidrule(lr){2-13}\cmidrule(lr){14-21}
      &
        \multicolumn{2}{c}{Area}
        & \multicolumn{2}{c}{Bar}
        & \multicolumn{2}{c}{Line}
        & \multicolumn{2}{c}{Pie}
        & \multicolumn{2}{c}{Scatter}
        & \multicolumn{2}{c}{All}
        & \multicolumn{2}{c}{Area}
        & \multicolumn{2}{c}{Bar}
        & \multicolumn{2}{c}{Line}
        & \multicolumn{2}{c}{All} \\
      &
          AS-l & UE-o
        & AS-l & UE-o
        & AS-l & UE-o
        & AS-l & UE-o
        & AS-l & UE-o
        & AS-l & UE-o
        & AS-l & UE-o
        & AS-l & UE-o
        & R-1 & R-L
        & R-1 & R-L \\
      \midrule
      \textbf{deepseek-vl2-tiny} \\

 Title
  & 13.17 & \textbf{79.00} & \textbf{26.04} & \textbf{76.55} & \textbf{16.35} & \textbf{78.61} & \textbf{27.69} & \textbf{81.54} 
  & \textbf{27.80} & \textbf{83.14} & \textbf{23.50} & \textbf{77.32} 
  & \textbf{7.27} & \textbf{84.34} & \textbf{5.30} & \textbf{78.45} & \textbf{7.78} & \textbf{84.36} & \textbf{6.43} & \textbf{81.47} \\
  
 Dict+Title
  & 4.80 & 53.55 & 3.29 & 47.27 & 4.03 & 51.71 & 26.52 & 66.96
  & 7.82 & 52.20 & 4.23 & 49.11 
  & 3.52 & 65.44 & 3.34 & 60.25 & 5.24 & 62.13 & 3.85 & 62.10 \\

 Statis+Title
  & 4.66 & 59.60 & 14.10 & 61.36 & 11.79 & 58.88 & 11.89 & 64.40
  & 7.48 & 58.17 & 13.25 & 60.77 
  & 2.49 & 67.19 & 2.47 & 63.54 & 2.87 & 64.39 & 2.57 & 64.73 \\

 Dict+Statis+Title
  & \textbf{15.80} & 51.09 & 16.00 & 55.53 & 14.61 & 52.43 & 13.97 & 56.76 
  & 5.63 & 53.43 & 15.47 & 54.70
  & 3.76 & 58.44 & 3.30 & 57.04 & 2.85 & 56.02 & 3.31 & 57.17 \\

 Dict+StatisT+Title
  & 6.64 & 60.28 & $/$ & $/$ & $/$ & $/$ & 24.54 & 65.79                   & 12.99  & 49.84 &  $/$ & $/$
  & 3.59 & 63.63 & 3.72 & 62.40 & 4.91 & 64.81 & 3.97 & 63.32 \\

\midrule
\textbf{internVL-2.5} \\

 Title
  & 12.02 & 75.29 & 25.30 & 77.44 & 19.36 & 78.32 & 27.34 & 81.97 
  & 13.18 & 82.42 & 23.55 & 77.82 
  & 6.15 & 84.00 & 5.71 & 82.39 & 8.52 & 83.62 & 6.51 & 83.12 \\

 Dict+Title
  & 33.85 & 81.43 & \textbf{39.15} & 85.36 & 32.86 & 83.51 & 33.17 & \textbf{83.75}
  & 33.14 & 86.08 & \textbf{37.26} & 84.80 
  & \textbf{11.00} & 86.38 & 6.01 & 84.41 & \textbf{14.31} & \textbf{86.88} & \textbf{9.36} & 85.54 \\

 Statis+Title
  & 23.73 & 80.98 & 38.08 & 84.03 & 34.56 & 83.07 & 24.64 & 78.52
  & \textbf{38.70} & \textbf{88.01} & 36.60 & 83.63 
  & 8.75 & 86.12 & 7.50 & 86.45 & 11.83 & 86.58 & 8.88 & 86.39 \\
  
 Dict+Statis+Title
  & \textbf{25.91} & 78.48 & 37.60 & 84.26 & \textbf{36.74} & \textbf{83.73} & 27.96 & 81.63 
  & 31.95 & 87.96 & 36.87 & 84.01
  & 10.79 & \textbf{86.60} & \textbf{7.71} & \textbf{86.55} & 10.76 & 86.06 & 9.28 & \textbf{86.44} \\

 Dict+StatisT+Title
  & 31.44 & \textbf{81.63} & 38.25 & \textbf{85.56} & 28.59 & 83.33 & \textbf{36.54} & 83.73  & 30.28 & 84.41 & 35.59 & \textbf{84.88}
  & 10.09 & 86.47 & 5.87 & 85.47 & 12.05 & 86.54 & 8.50 & 86.00 \\

\midrule
\textbf{qwen2.5-VL-3B} \\
 Title
  & 19.93 & 80.57 & \textbf{35.17} & \textbf{84.30} & 23.08 & \textbf{81.29} & 37.93 & \textbf{87.19} 
  & 23.09 & \textbf{90.38} & \textbf{31.87} & \textbf{83.64} 
  & 7.74 & \textbf{84.92} & \textbf{6.75} & \textbf{82.17} & 10.64 & \textbf{84.38} & 7.96 & \textbf{83.45} \\
  
 Dict+Title
  & \textbf{31.04} & 78.27 & 31.97 & 80.16 & 25.50 & 80.53 & 40.41 & 86.07
  & \textbf{31.48} & 87.28 & 30.58 & 80.47
  & 9.72 & 81.90 & \textbf{6.75} & 79.62 & 11.06 & 81.67 & 8.59 & 80.73 \\

 Statis+Title
  & 26.80 & \textbf{81.39} & 32.09 & 79.64 & \textbf{26.93} & 78.87 & \textbf{40.72} & 85.56 
  & 24.60 & 86.61 & 30.89 & 79.72 
  & \textbf{10.99} & 82.32 & 6.60 & 80.58 & \textbf{13.65} & 82.60 & \textbf{9.49} & 81.54 \\

 Dict+Statis+Title
  & 23.41 & 79.91 & 33.29 & 80.53 & 26.66 & 80.84 & 35.81 & 85.10 
  & 26.57 & 86.92 & 31.49 & 80.80 
  & 10.89 & 82.51 & 5.47 & 80.11 & 12.92 & 83.60 & 8.73 & 81.60 \\

 Dict+StatisT+Title
  & 17.21 & 76.84 & 32.93 & 80.63 & 26.62 & 80.72 & 33.39 & 86.96
  & 24.70 & 81.14 & 31.05 & 80.79 
  & 10.73 & 80.86 & 6.01 & 79.77 & 12.89 & 83.94 & 8.95 & 81.07 \\

      \bottomrule
    \end{tabular}%
  }
  \caption{Ablation study results (\textbf{A}lignScore-\textbf{l}arge / \textbf{U}niEval-\textbf{o}verall) for different models regarding data used on Pew and VisText datasets.}
  \label{tab:apix_ablation_textpye_aliuni}
\end{table*}

\begin{table*}[t]
  \centering
  \renewcommand{\arraystretch}{1.2}
  \resizebox{\textwidth}{!}{%
    \begin{tabular}{ll*{20}{r}}
      \toprule
      \multirow{3}{*}{\shortstack{\textbf{VLM} \\  \\ \textbf{+Textual Data}}}
        & \multicolumn{12}{c}{\textbf{Pew}}
        & \multicolumn{8}{c}{\textbf{VisText}} \\
      \cmidrule(lr){2-13}\cmidrule(lr){14-21}
      &
        \multicolumn{2}{c}{Area}
        & \multicolumn{2}{c}{Bar}
        & \multicolumn{2}{c}{Line}
        & \multicolumn{2}{c}{Pie}
        & \multicolumn{2}{c}{Scatter}
        & \multicolumn{2}{c}{All}
        & \multicolumn{2}{c}{Area}
        & \multicolumn{2}{c}{Bar}
        & \multicolumn{2}{c}{Line}
        & \multicolumn{2}{c}{All} \\
      &
          coh & cons
        & coh & cons
        & coh & cons
        & coh & cons
        & coh & cons
        & coh & cons
        & coh & cons
        & coh & cons
        & coh & cons
        & coh & cons \\
      \midrule
      \textbf{deepseek-vl2-tiny} \\

 Title
  & \textbf{86.91} & \textbf{48.42} & \textbf{80.57} & \textbf{55.94} & \textbf{85.51} & \textbf{52.75} & \textbf{85.14} & \textbf{65.36} 
  & \textbf{84.82} & \textbf{71.25} & \textbf{82.08} & \textbf{55.48}
  & \textbf{87.91} & \textbf{69.66} & \textbf{82.15} & \textbf{60.05} & \textbf{87.84} & \textbf{69.94} & \textbf{85.08} & \textbf{65.03} \\
  
 Dict+Title
  & 44.42 & 46.63 & 38.93 & 33.56 & 43.52 & 40.27 & 62.19 & 62.99
  & 43.02 & 45.05 & 40.89 & 36.42 
  & 63.46 & 66.28 & 57.19 & 56.75 & 59.39 & 63.18 & 59.41 & 60.88 \\

 Statis+Title
  & 51.95 & 46.01 & 54.43 & 53.74 & 51.70 & 50.76 & 60.55 & 54.46
  & 48.75 & 49.90 & 53.82 & 52.86 
  & 64.43 & 64.64 & 59.53 & 57.99 & 61.14 & 60.35 & 61.24 & 60.35 \\

 Dict+Statis+Title
  & 55.53 & 35.87 & 56.77 & 49.53 & 55.20 & 37.07 & 54.40 & 50.30 
  & 58.58 & 51.30 & 56.31 & 46.25 
  & 66.74 & 55.07 & 63.70 & 51.76 & 63.63 & 53.65 & 64.50 & 53.11 \\

 Dict+StatisT+Title
  & 59.99 & 32.63 & $/$ & $/$ & $/$ & $/$ & 59.90 & 58.31                   & 46.11  & 45.69 &  $/$ & $/$
  & 60.57 & 51.09 & 59.99 & 50.98 & 62.62 & 51.49 & 60.78 & 51.14 \\

\midrule
\textbf{internVL-2.5} \\

 Title
  & 81.15 & \textbf{55.46} & 81.70 & 60.16 & 84.77 & 55.98 & 88.64 & \textbf{67.23}
  & 89.11 & 59.54 & 82.74 & 59.24 
  & 90.54 & 62.16 & 88.90 & 60.34 & 89.56 & 62.71 & 89.50 & 61.41 \\

 Dict+Title
  & 90.89 & 49.32 & 93.20 & 61.16 & 92.01 & 56.10 & 89.73 & 62.28 
  & 91.65 & 66.91 & 92.75 & 59.81 
  & \textbf{93.29} & 64.94 & 93.18 & 58.34 & 93.59 & 66.25 & 93.31 & 62.03 \\

 Statis+Title
  & 90.60 & 49.78 & 90.00 & 62.09 & 89.55 & \textbf{59.69} & 82.30 & 56.17 
  & 93.64 & \textbf{70.09} & 89.71 & \textbf{61.23} 
  & 93.19 & 63.93 & \textbf{93.68} & \textbf{64.66} & \textbf{93.61} & 64.87 & \textbf{93.53} & \textbf{64.51} \\
  
 Dict+Statis+Title
  & 89.87 & 47.52 & 93.15 & \textbf{62.10} & \textbf{93.01} & 54.95 & \textbf{93.07} & 65.21 
  & \textbf{95.84} & 68.54 & \textbf{93.09} & 60.26 
  & 93.07 & 65.42 & 93.12 & 61.82 & 93.38 & 66.56 & 93.17 & 63.93 \\
  
 Dict+StatisT+Title
  & \textbf{91.59} & 49.14 & \textbf{93.42} & 61.72 & 92.44 & 54.33 & 89.19 & 64.25 
  & 91.37 & 60.36 & 93.00 & 59.75 
  & 92.73 & \textbf{66.21} & 93.33 & 62.25 & 92.75 & \textbf{66.78} & 93.03 & 64.41 \\

\midrule
\textbf{qwen2.5-VL-3B} \\
 Title
  & \textbf{87.39} & 53.39 & \textbf{88.87} & \textbf{66.98} & \textbf{87.14} & 58.37 & \textbf{91.00} & 72.48 
  & \textbf{94.42} & \textbf{78.91} & \textbf{88.54} & \textbf{64.92} 
  & \textbf{87.62} & \textbf{72.22} & \textbf{86.76} & \textbf{64.40} & \textbf{87.17} & 71.31 & \textbf{87.09} & \textbf{68.18} \\

 Dict+Title
  & 81.76 & 55.36 & 83.62 & 61.79 & 85.71 & \textbf{58.47} & 86.86 & 77.68 
  & 91.17 & 73.52 & 84.30 & 61.46 
  & 84.63 & 67.65 & 83.51 & 61.10 & 83.93 & 67.96 & 83.91 & 64.53 \\
  
 Statis+Title
  & 87.38 & \textbf{57.06} & 82.68 & 61.71 & 83.35 & 56.77 & 87.48 & 73.66 
  & 93.30 & 65.71 & 83.17 & 60.80 
  & 85.33 & 67.24 & 84.06 & 63.47 & 85.32 & 68.70 & 84.71 & 65.75 \\
  
 Dict+Statis+Title
  & 87.18 & 51.65 & 84.73 & 60.84 & 86.28 & 57.91 & 85.89 & 76.50 
  & 92.69 & 68.51 & 85.28 & 60.51 
  & 85.63 & 67.22 & 84.26 & 61.95 & 86.83 & 69.25 & 85.25 & 65.14 \\

 Dict+StatisT+Title
  & 80.49 & 52.85 & 84.69 & 61.54 & 85.87 & 58.36 & 88.08 & \textbf{78.51}
  & 82.55 & 65.05 & 85.01 & 61.16 
  & 83.15 & 67.21 & 84.35 & 60.77 & 86.46 & \textbf{71.47} & 84.53 & 65.09 \\

      \bottomrule
    \end{tabular}%
  }
  \caption{Ablation study results (UniEval-\textbf{coh}erence / UniEval-\textbf{cons}istency) for different models regarding data used on Pew and VisText datasets.}
  \label{tab:apix_ablation_textpye_aliunicohcon}
\end{table*}

\begin{table*}[t]
  \centering
  \renewcommand{\arraystretch}{1.2}
  \resizebox{\textwidth}{!}{%
    \begin{tabular}{ll*{20}{r}}
      \toprule
      \multirow{3}{*}{\shortstack{\textbf{VLM} \\  \\ \textbf{+Textual Data}}}
        & \multicolumn{12}{c}{\textbf{Pew}}
        & \multicolumn{8}{c}{\textbf{VisText}} \\
      \cmidrule(lr){2-13}\cmidrule(lr){14-21}
      &
        \multicolumn{2}{c}{Area}
        & \multicolumn{2}{c}{Bar}
        & \multicolumn{2}{c}{Line}
        & \multicolumn{2}{c}{Pie}
        & \multicolumn{2}{c}{Scatter}
        & \multicolumn{2}{c}{All}
        & \multicolumn{2}{c}{Area}
        & \multicolumn{2}{c}{Bar}
        & \multicolumn{2}{c}{Line}
        & \multicolumn{2}{c}{All} \\
      &
          flu & rel
        & flu & rel
        & flu & rel
        & flu & rel
        & flu & rel
        & flu & rel
        & flu & rel
        & flu & rel
        & flu & rel
        & flu & rel \\
      \midrule
      \textbf{deepseek-vl2-tiny} \\

 Title
  & \textbf{95.08} & \textbf{85.59} & \textbf{90.91} & \textbf{78.78} & \textbf{92.45} & \textbf{83.72} & \textbf{91.45} & \textbf{84.22} 
  & \textbf{92.99} & \textbf{83.48} & \textbf{91.40} & \textbf{80.33} 
  & \textbf{94.63} & \textbf{85.16} & \textbf{92.39} & \textbf{79.22} & \textbf{94.33} & \textbf{85.33} & \textbf{93.47} & \textbf{82.30} \\
  
 Dict+Title
  & 81.05 & 42.11 & 85.37 & 31.23 & 85.85 & 37.21 & 83.11 & 59.55
  & 79.72 & 41.02 & 85.30 & 33.82 
  & 72.96 & 59.07 & 75.20 & 51.85 & 71.72 & 54.22 & 73.76 & 54.37 \\

 Statis+Title
  & 90.66 & 49.79 & 86.26 & 51.00 & 84.58 & 48.49 & 86.54 & 56.07
  & 90.35 & 43.67 & 85.96 & 50.42 
  & 79.90 & 59.77 & 83.34 & 53.30 & 80.20 & 55.86 & 81.66 & 55.66 \\

 Dict+Statis+Title
  & 92.76 & 53.44 & 85.63 & 55.13 & 86.96 & 53.77 & 83.35 & 52.61 
  & 83.83 & 55.49 & 85.98 & 54.69
  & 88.18 & 63.66 & 87.31 & 58.60 & 87.48 & 59.61 & 87.59 & 60.21 \\

 Dict+StatisT+Title
  & 91.16 & 57.35 & $/$ & $/$ & $/$ & $/$ & 86.83 & 58.14                   & 65.16  & 42.38 &  $/$ & $/$
  & 85.35 & 57.51 & 83.39 & 55.26 & 86.14 & 58.99 & 84.58 & 56.77 \\

\midrule
\textbf{internVL-2.5} \\

 Title
  & 83.32 & 81.21 & 86.38 & 81.52 & 87.87 & 84.65 & 84.43 & 87.57 
  & 91.29 & 89.71 & 86.71 & 82.57 
  & 93.57 & 89.73 & 92.17 & 88.12 & 92.86 & 89.33 & 92.72 & 88.85 \\

 Dict+Title
  & \textbf{95.01} & 90.50 & 94.25 & 92.83 & 94.27 & 91.66 & 94.59 & 88.39 
  & 94.88 & 90.88 & 94.29 & 92.35 
  & 95.10 & 92.18 & 94.12 & 92.02 & 95.02 & \textbf{92.67} & 94.60 & 92.22 \\

 Statis+Title
  & 94.06 & 89.51 & \textbf{94.81} & 89.23 & 94.48 & 88.57 & 95.25 & 80.34
  & \textbf{95.82} & 92.48 & \textbf{94.74} & 88.84 
  & \textbf{95.30} & 92.04 & \textbf{95.15} & \textbf{92.30} & \textbf{95.43} & 92.40 & \textbf{95.26} & \textbf{92.26} \\
  
 Dict+Statis+Title
  & 94.65 & 88.82 & 94.61 & 92.83 & 94.25 & \textbf{92.67} & \textbf{95.41} & \textbf{92.24} 
  & 95.38 & \textbf{95.25} & 94.56 & \textbf{92.74}
  & 95.18 & \textbf{92.26} & 94.26 & 92.08 & 95.07 & 92.33 & 94.71 & 92.19 \\

 Dict+StatisT+Title
  & 94.53 & \textbf{91.27} & 94.17 & \textbf{92.95} & \textbf{94.60} & 91.97 & 93.76 & 87.71 
  & 94.88 & 91.03 & 94.28 & 92.50 
  & 95.27 & 91.68 & 94.08 & 92.22 & 95.01 & 91.60 & 94.62 & 91.93 \\

\midrule
\textbf{qwen2.5-VL-3B} \\
 Title
  & \textbf{95.15} & 86.34 & \textbf{93.70} & \textbf{87.65} & \textbf{93.40} & \textbf{86.26} & \textbf{95.07} & \textbf{90.19} 
  & \textbf{95.46} & 92.74 & \textbf{93.71} & \textbf{87.41} 
  & \textbf{95.06} & \textbf{84.77} & \textbf{93.11} & \textbf{84.42} & \textbf{94.73} & 84.32 & \textbf{94.03} & \textbf{84.49} \\

 Dict+Title
  & 94.04 & 81.93 & 91.98 & 83.23 & 92.55 & 85.36 & 93.42 & 86.32 
  & 93.73 & 90.72 & 92.21 & 83.92 
  & 93.35 & 81.98 & 92.87 & 81.03 & 93.77 & 81.02 & 93.21 & 81.28 \\
  
 Statis+Title
  & 93.91 & \textbf{87.23} & 92.16 & 82.01 & 92.50 & 82.87 & 93.78 & 87.31 
  & 94.45 & \textbf{92.99} & 92.34 & 82.58 
  & 93.69 & 83.03 & 93.08 & 81.69 & 93.85 & 82.52 & 93.43 & 82.25 \\

 Dict+Statis+Title
  & 94.14 & 86.65 & 92.30 & 84.26 & 93.21 & 85.97 & 92.69 & 85.33 
  & 94.22 & 92.27 & 92.59 & 84.84 
  & 94.14 & 83.06 & 92.48 & 81.76 & 93.92 & \textbf{84.41} & 93.27 & 82.75 \\
  
 Dict+StatisT+Title
  & 93.59 & 80.41 & 92.02 & 84.25 & 93.27 & 85.37 & 93.79 & 87.47
  & 93.50 & 83.47 & 92.42 & 84.56 
  & 92.85 & 80.23 & 92.15 & 81.81 & 93.82 & 84.01 & 92.74 & 81.91 \\
      \bottomrule
    \end{tabular}%
  }
  \caption{Ablation study results (UniEval-\textbf{flu}ency / UniEval-\textbf{rel}evance) for different models regarding data used on Pew and VisText datasets.}
  \label{tab:apix_ablation_textpye_aliuniflurel}
\end{table*}

\section{Case Study}
\label{sec:apix_case_study}
\begin{figure*}[h]
    \centering
    \includegraphics[width=0.98\linewidth]{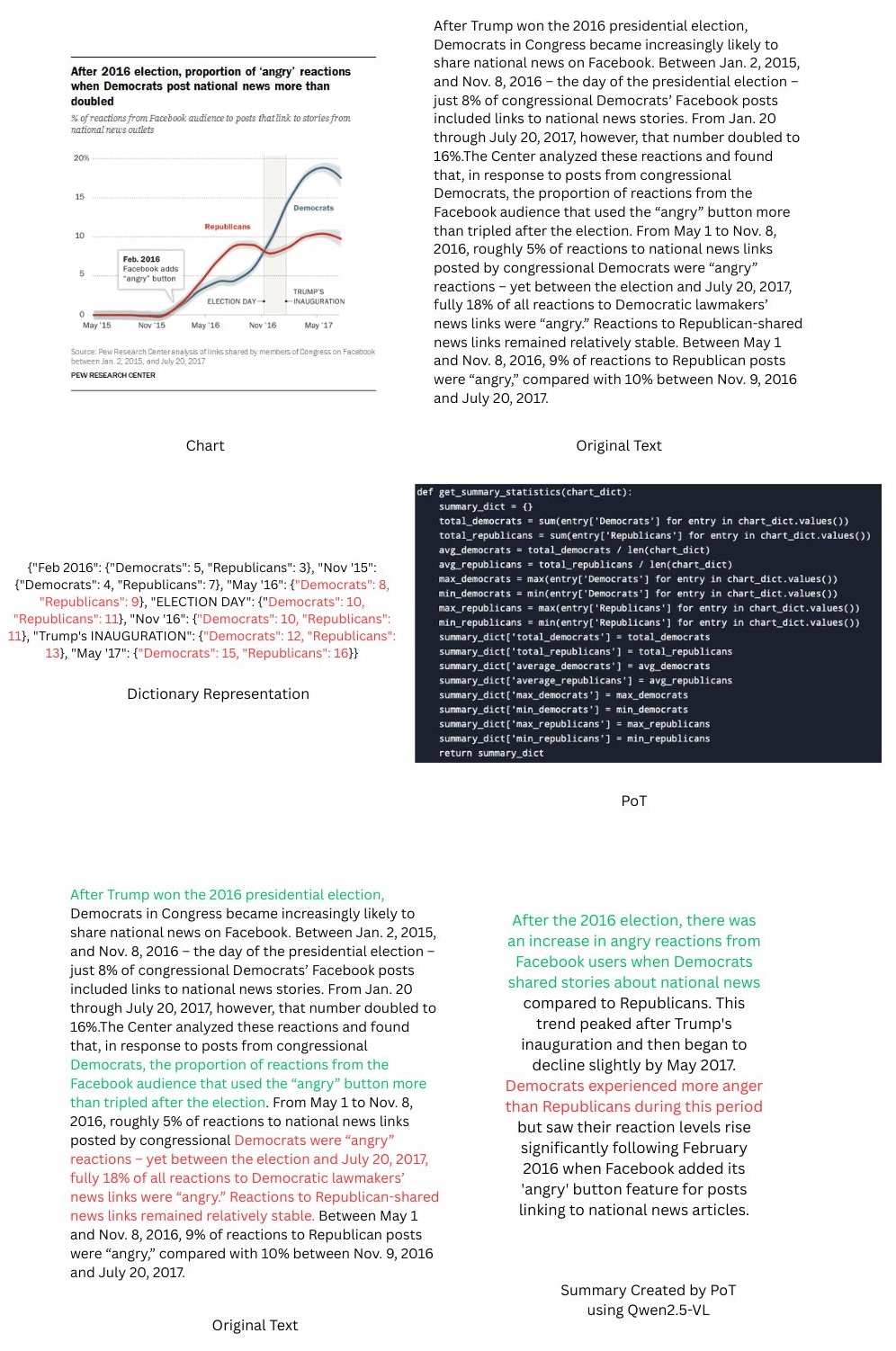}
    \caption{Case study on the generated dictionary, PoT, and generated caption from the experiment trials.}
    \label{fig:case_study}
\end{figure*}

A case study in Figure~\ref{fig:case_study} demonstrates an end-to-end chart-to-text method using the PoT. 
In this specific instance, the chart-to-dictionary properly captures the appropriate format of how to organize the data, but fundamentally mislabels or misreads the values of which values go to which parties. However, it can be observed that in terms of observing the increasing trend in the time-series data, the dictionary was able to somewhat capture this. 
The generated PoT is agnostic of the actual values of the functions and is able to correctly identify the relevant keys needed to create summary statistics of total, average, and min and max values. 
The generated caption captures the general ideas that the chart was able to portray, specifically describing the chart elements of date in the x-axis and anger in the y-axis. While not as verbose as the original text, the generated summary was able to capture the key ideas and trends in the caption.

\section{Failure Case Analysis}
\label{sec:apix_failure_case_study}

\subsection{Python Dictionary Generation}

In order to keep the desired quality of the statistics in this work, we decided to use InternVL-2.5-4B \cite{chen2024internvl} with ChatGPT-4o-mini \cite{openai2024gpt4omini} to generate the data dictionary.
Figure~\ref{fig:dict_failurenum} shows comparisons of failure numbers of the chart data dictionary generation by each VLM, presenting InternVL has the best capability on handling and generating more data dictionaries from the chart data.
Since LLaVA is primarily an LLM (LLaMA) with a vision adapter, whereas DeepSeek, InternVL, and Qwen are specialized vision-language models with strong visual encoding, we test DeepSeek, InternVL, and Qwen on generating the dictionary for chart data on Pew and VisText datasets, respectively.

But we are aware that most failure cases are due to (1) limitation on maximum LLM output length, so the output Python code is cut off a part; (2) complex structure or format of the JSON data or Python style, which cannot be generally read, recognized, or pass the execution tests, and are consequently categorized as failure cases, rather than nonsense or empty outputs. Common error message instances are collected and listed in Table~\ref{tab:err_message_pydict}. In future work, we will implement a module to refine the Python code into their correct format, ensuring the collection of all valuable data.

\begin{table*}[h!]
\centering
\begin{tabular}{l}
\toprule
\textbf{Too long output to be cut off a part} \\
\midrule
\texttt{'$[$' was never closed (<string>, line 40)} \\
\texttt{'$\{$' was never closed (<string>, line 74)} \\
\midrule
\textbf{Complex dictionary data structures to be read} \\
\midrule
\texttt{unsupported operand type(s) for +: 'int' and 'str'}       \\
\texttt{unsupported operand type(s) for +: 'int' and 'list'}      \\
\texttt{unsupported operand type(s) for +: 'int' and 'dict'}      \\
\texttt{unsupported operand type(s) for +: 'int' and 'NoneType'}  \\
\midrule
\textbf{Other specific data to be read} \\
\midrule
\texttt{'int' object has no attribute 'values'} \\
\texttt{invalid literal for int() with base 10: '\$30K-\$99999'} \\
\texttt{unterminated string literal (detected at line 24) (<string>, line 24)} \\
\texttt{unterminated f-string literal (detected at line 44) (<string>, line 44)} \\
\bottomrule
\end{tabular}
\caption{Error message instances from Python dictionary generation failure cases.}
\label{tab:err_message_pydict}
\end{table*}

\subsection{Python Code Generation}
\begin{figure*}[h]
  \includegraphics[width=\linewidth]{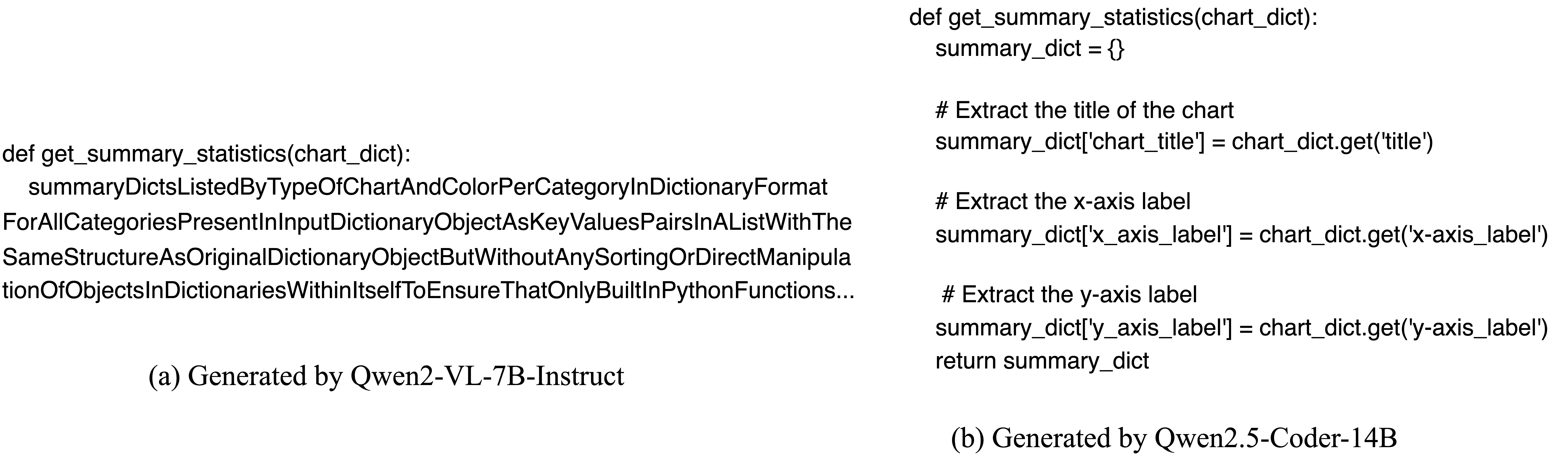}
  \caption{Comparison of failed generated Python code by the general-purpose LLM and the desired generated Python code by the code-specialized LLM.}
  \label{fig:python_codes}
\end{figure*}

Figure~\ref{fig:python_codes} presents a comparison between the failure-prone code generated by general-purpose LLMs and the acceptable code produced by code-specialized LLMs, where those models were specifically pre-trained and fine-tuned on programming codes, such as Qwen-Coder.
With this observation, we chose to use Qwen-2.5-Coder-14B \cite{hui2024qwen25coder}, which is optimized for generating accurate and efficient code outputs, to ensure the quality of the generated code.

\section{Human Evaluation Details}
\label{sec:apix_human_evaluation}

\begin{figure*}[h]
    \centering
    \includegraphics[width=0.98\linewidth]{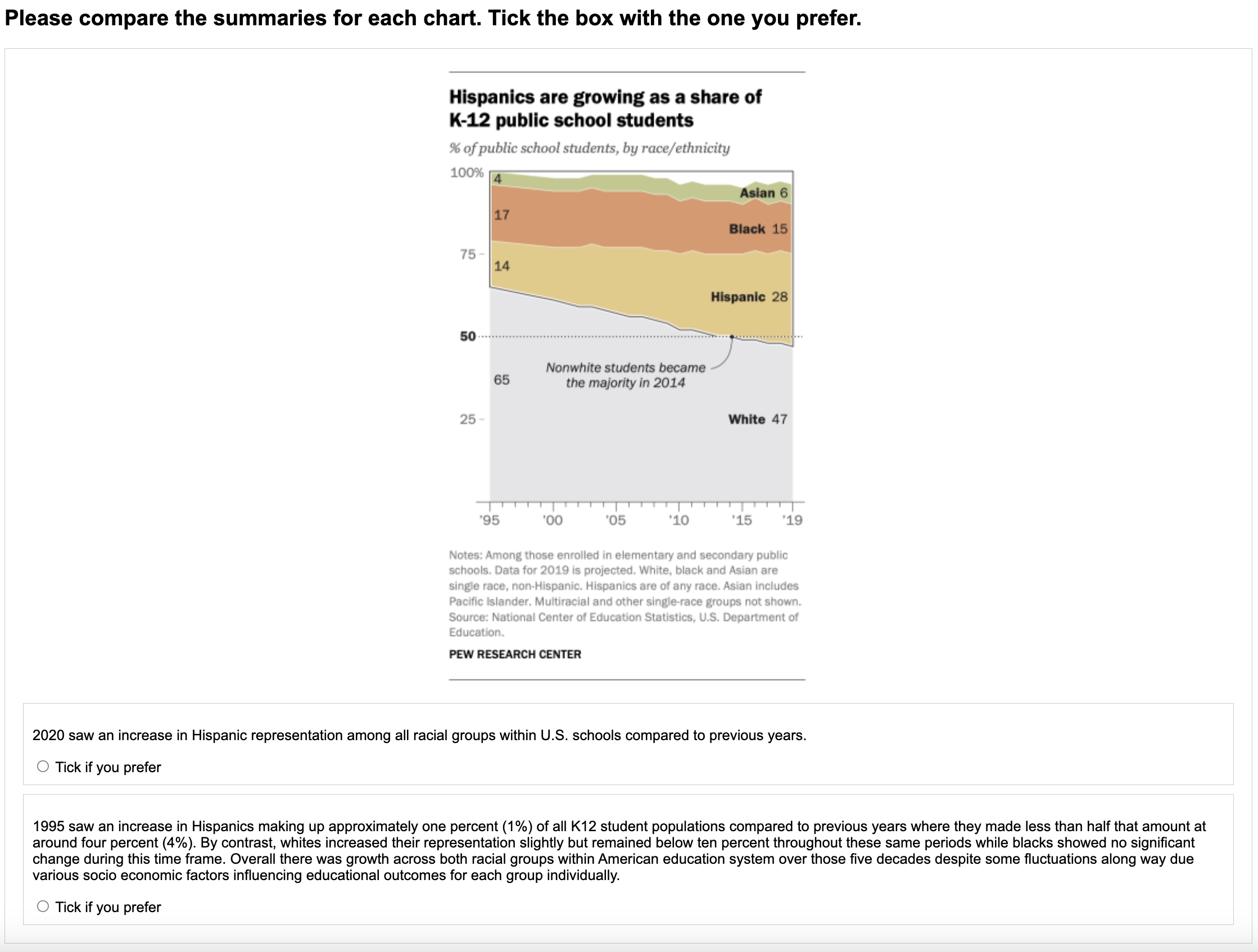}
    \caption{An exemplary screenshot of an instance for human evaluation on our webpage.}
    \label{fig:website_screenshot}
\end{figure*}

We randomly selected 50 chart samples from both datasets, comprising outputs from both the template-based and PoT-based methods, with 10 charts sampled for each chart type across area, bar, line, pie, and scatter charts. Three graduate students (Master’s and PhD) who research in natural language processing were invited for the human evaluation as volunteers. As participation was voluntary, no payment-related considerations apply. We developed a webpage where human evaluators were requested to select their preferred summary for each pair of summaries of the provided chart. The average total number of selections of each summary category serves as its human evaluation score. An exemplary screenshot of an instance shown on our webpage is shown in Figure~\ref{fig:website_screenshot}. The evaluation results indicate the effectiveness of using PoT in generating statistics content for improving the summary quality in chart summarization compared to rule-based statistics extraction.

\end{document}